\documentclass{article}

\usepackage[english]{babel}

\usepackage[letterpaper,top=2cm,bottom=2cm,left=3cm,right=3cm,marginparwidth=1.75cm]{geometry}

\usepackage{setspace}
\usepackage[utf8]{inputenc} 
\usepackage[T1]{fontenc}    
\usepackage{url}            
\usepackage{booktabs}       
\usepackage{amsmath}
\usepackage{amsfonts}       
\usepackage{nicefrac}       
\usepackage{xcolor}         
\usepackage{amssymb}
\usepackage{tikz}
\usepackage{comment}
\usepackage{pifont}
\usepackage{bbm}
\usepackage{multirow}
\usepackage{rotating}
\usepackage{graphicx}
\usepackage{colortbl}       
\usepackage{wrapfig}
\usepackage[colorlinks=true, allcolors={blue!55!black}]{hyperref}
\usepackage{authblk}
\usepackage{placeins}   
\usepackage[numbers,sort&compress]{natbib}
\usepackage[final]{microtype}
\usepackage{booktabs}
\usepackage{longtable}
\usepackage{pdflscape}
\usepackage{threeparttable}
\usepackage{caption}
\usepackage{pdfpages}
\usepackage{graphicx}
\usepackage{pgffor}
\usepackage{pdftexcmds}
\usepackage{lineno}
\usepackage{fvextra}
\usepackage{newunicodechar}
\newunicodechar{☆}{\ensuremath{\star}}
\newcommand{\framedpdfpage}[2]{
  \thispagestyle{plain}%
  \begin{center}
    \setlength{\fboxsep}{5pt}%
    \setlength{\fboxrule}{0.4pt}%
    \fbox{\includegraphics[page=#2, height=0.86\textheight, keepaspectratio]{#1}}%
  \end{center}
  \clearpage}

\renewenvironment{abstract}{%
  \small
  \begin{center}{\bfseries \abstractname\vspace{-.5em}}\end{center}%
  \quote
}{%
  \endquote
}

\ifdefined\CBox\else\newsavebox\CBox\fi

\date{}

\title{\textbf{LUCAID: Agentic Multimodal AI for Lung Cancer Precision Pathology}}
\renewcommand*{\thefootnote}{\fnsymbol{footnote}}

\author[1,2,\#]{Marie-Lisa Eich}
\author[3,\#]{Kai Standvoss}
\author[3,\#]{Timo Milbich}
\author[3,6,10,\#]{Alexander M\"ollers}
\author[3]{Miriam H\"agele}
\author[4]{Philipp Anders}
\author[4]{Lars Tharun}
\author[3]{Hanna Kontradiuk}
\author[3]{Sebastian Kons}
\author[3]{Nader Aldoj}
\author[3]{Recepcan Adig\"uzel}
\author[3]{Adam Narai}
\author[3]{Lukas H\"onig}
\author[3]{Jonathan Striebel}
\author[1]{Binru Yang}
\author[1,13]{Mihnea P. Dragomir}
\author[3,6,10]{Marvin Sextro}
\author[1,10,11]{Philipp Keyl}
\author[11]{Philipp Jurmeister}
\author[3]{Rosemarie Krupar}
\author[3]{Evelyn Ramberger}
\author[3]{James Wells}
\author[3]{Julika Ribbat-Idel}
\author[3]{Andreas Kunft}
\author[5]{Hussam Shuaib}
\author[5]{Christian Groh\'e}
\author[15]{Reinhard B\"uttner}
\author[1]{David Horst}
\author[6,7,8,9,10]{Klaus-Robert M\"uller}
\author[3]{Lukas Ruff}
\author[3,+]{Maximilian Alber}
\author[1,3,10,11,12,14,+]{Frederick Klauschen}
\author[1,13,+]{Simon Schallenberg}

\affil[1]{Institute of Pathology, Charit\'e -- Universit\"atsmedizin Berlin, corporate member of Freie Universit\"at Berlin, Humboldt-Universit\"at zu Berlin, Germany}
\affil[2]{Berlin Institute of Health at Charit\'e -- Universit\"atsmedizin Berlin, BIH Biomedical Innovation Academy, BIH Charit\'e Digital Clinician Scientist Program, Charit\'eplatz 1, 10117 Berlin, Germany}
\affil[3]{Aignostics GmbH, Berlin, Germany}
\affil[4]{MVZ HPH Institut f\"ur Pathologie und H\"amatopathologie GmbH, Hamburg, Germany}
\affil[5]{Evangelische Lungenklinik Berlin-Buch, Berlin, Germany}
\affil[6]{Machine Learning Group, Technical University of Berlin, Berlin, Germany}
\affil[7]{Department of Mathematics and Computer Science, Technical University of Berlin, Germany}
\affil[8]{Department of Artificial Intelligence, Korea University, Seoul 136-713, South Korea}
\affil[9]{MPI for Informatics, Saarbr\"ucken, Germany}
\affil[10]{BIFOLD -- Berlin Institute for the Foundations of Learning and Data, Berlin, Germany}
\affil[11]{Institute of Pathology, Ludwig-Maximilians-University, Munich, Germany}
\affil[12]{German Cancer Consortium (DKTK), German Cancer Research Center (DKFZ), Munich Partner Site, Heidelberg, Germany}
\affil[13]{German Cancer Consortium (DKTK), German Cancer Research Center (DKFZ), Berlin Partner Site, Heidelberg, Germany}
\affil[14]{Bavarian Cancer Research Center (BZKF), Munich Partner Site, Munich, Germany}
\affil[15]{Institute of Pathology, University Hospital Cologne, Cologne, Germany}

\usepackage{fancyhdr}       
\fancypagestyle{plain}{
  \fancyhf{}%
  \fancyfoot[R]{\thepage\hspace*{-1cm}}%
}

\begin{document}
\maketitle
\thispagestyle{empty}
\vspace{-3em}
\begin{abstract}

Lung cancer tissue diagnostics is complex, as therapy decisions in precision oncology rely on the integration of histomorphological, immunohistochemical, and molecular features. Yet pathological assessment remains largely visual and semi-quantitative and shows interobserver variability, while existing artificial intelligence (AI) tools cover only selected tasks, rarely reach generalizable expert-level performance, and lack prospective clinical validation. To address these challenges, we developed and clinically validated LUCAID, an agentic AI system for precision lung cancer pathology. An integrative agent couples diagnostic reasoning with nine modules that cover the full routine workflow, from quality control, tumor detection and segmentation, histological subtyping, tumor microenvironment profiling, tumor cellularity quantification, and predictive biomarker scoring (PD-L1, MET, TROP-2) to automated structured report generation. LUCAID enables users to interactively query the module outputs and generate reports that contextualize the results. Against large-scale expert ground-truth annotations, the analysis modules achieved F1 scores of 0.82–0.95. In prospective clinical validation, LUCAID reached 93.0\% concordance with an expert-panel adjudicated reference standard across clinically actionable decisions, compared with 68.3–81.1\% for five experienced thoracic pathologists.

\end{abstract}

\renewcommand\thefootnote{\#}\footnotetext{Contributed equally to this work}
\renewcommand\thefootnote{+}\footnotetext{Contributed equally to this work}

\begin{figure}[p]
    \vspace{-25mm}
    \centering
    \includegraphics[width=\linewidth,height=0.7\textheight,keepaspectratio,trim={0 4.5cm 0 0},clip]{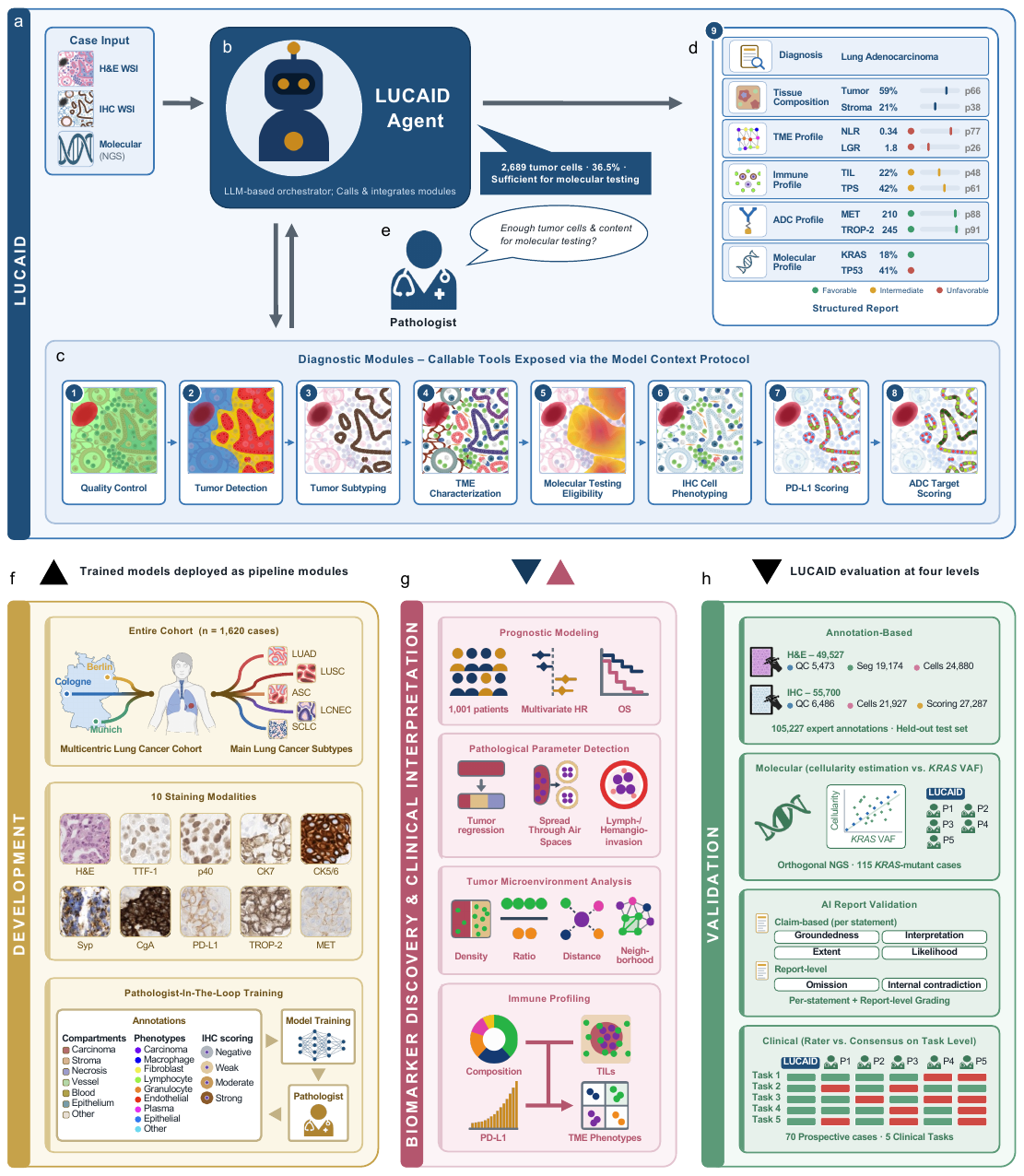}
    \small
    \caption[Workflow for LUCAID.]{\textbf{LUCAID system overview, development, biomarker discovery and validation.}\newline
    \textbf{a},~Case input: digitized H\&E and IHC 
whole-slide images  from a routine lung cancer case, together with
molecular profiling results, enter the system.
\textbf{b},~A large language model (LLM)-based agent coordinates the
analysis by calling diagnostic modules as needed and integrating their
validated quantitative results.
\textbf{c},~Analysis modules (left to right): (1)~quality control (QC);
(2)~tumor detection via tissue segmentation; (3)~tumor subtyping;
(4)~tumor microenvironment (TME) characterization; (5)~tumor cellularity
assessment; (6)~IHC cell phenotyping; (7)~PD-L1 scoring; and
(8)~antibody--drug conjugate (ADC) target expression scoring for TROP-2
and MET.
\textbf{d},~The agent integrates the module results and generates (9)~a
structured report comprising six sections: diagnosis, tissue composition,
TME profile, immune profile, ADC profile and molecular profile.
\textbf{e},~Throughout the analysis, the pathologist can pose
case-specific clinical queries in natural language and receive
quantitative, reproducible answers based on the module outputs.
\textbf{f},~Development: a multicentric cohort of 1{,}620 lung cancer
cases encompassing all major histological subtypes included digitized
H\&E and IHC WSIs spanning ten staining modalities. Models were
trained using a pathologist-in-the-loop approach with H\&E and IHC
annotations.
\textbf{g},~Biomarker discovery and clinical interpretation: LUCAID was
applied to a discovery cohort of 1{,}001 patients to quantify established
pathological features, investigate candidate biomarkers, characterize the
tumor and immune microenvironment, and contextualize individual patient
findings using cohort-level reference distributions.
\textbf{h},~Validation at four levels: annotation-based benchmarking using
105{,}227 expert annotations across H\&E and IHC; AI report validation
using claim- and report-level assessment of groundedness, interpretation,
extent, likelihood, omission and internal contradiction; molecular
validation against orthogonal next-generation sequencing data from
115~\textit{KRAS}-mutant cases; and prospective multicentric clinical
validation in 70 cases across five pathologists and five clinically
actionable tasks.\newline
\textit{Abbreviations:} ADC, antibody--drug conjugate; ASC, adenosquamous
carcinoma; CgA, chromogranin~A; CK5/6, cytokeratin 5/6; CK7, cytokeratin 7;
H\&E, hematoxylin and eosin; IHC, immunohistochemical; LCNEC, large cell
neuroendocrine carcinoma; LLM, large language model; LUAD, lung
adenocarcinoma; LUSC, lung squamous cell carcinoma; MET, hepatocyte growth
factor receptor; PD-L1, programmed death-ligand 1; QC, quality control;
SCLC, small cell lung cancer; SYP, synaptophysin; TME, tumor microenvironment;
TROP-2,  trophoblast cell-surface antigen 2; TTF-1, thyroid
transcription factor 1; WSI, whole-slide image.}
\label{fig1:workflow}
\end{figure}

\section{Introduction} \label{sec:intro}

Lung cancer remains the leading cause of cancer-related mortality worldwide, accounting for approximately 1.8 million deaths annually \citep{bray_global_2024}. Recent advances in targeted therapies, immune checkpoint inhibitors, and antibody-drug conjugates (ADCs) have substantially expanded treatment options for patients with actionable biomarkers \citep{howlader_effect_2020, reck_pembrolizumab_2016, wu_osimertinib_2020, felip_adjuvant_2021,kastner_evaluation_2024, passaro_esmo_2022, hanna_therapy_2021}. Consequently, lung cancer pathology has evolved from a predominantly morphological discipline into a multimodal field that increasingly depends on evaluating and integrating histopathological, immunohistochemical, and molecular information. Therapeutic decisions now require several distinct assessments: reliable identification of diagnostically evaluable tumor tissue, precise histological subtyping, quantitative assessment of subcellular biomarker expression, and integration of these measurements with molecular profiling results \citep{zer_early_2025, hendriks_oncogene-addicted_2023, camidge_telisotuzumab_2024, lawrence_fda_2025,ahn_datopotamab_2025, malone_molecular_2020, walsh_current_2024}. 

However, the rigorous and standardized quantification of these histopathological features remains challenging in the daily diagnostic workflow \citep{lami_overcoming_2023, robert_high_2023,volynskaya_ki67_2019}. Furthermore, novel candidate biomarkers within the tumor microenvironment (TME), such as the density and spatial distribution of tumor-infiltrating lymphocytes \citep{park_artificial_2022, yan_prognostic_2024}, are continuously emerging but are impractical to evaluate manually at scale. This complexity grows further with the immunohistochemical biomarkers required to guide therapeutic decisions, which are increasing in number and in the complexity of their assessment. The list of candidates is expanding rapidly beyond established targets such as programmed cell death-ligand 1 (PD-L1) and the ADC targets trophoblast cell-surface antigen 2 (TROP-2) and hepatocyte growth factor receptor (MET) \citep{lawrence_fda_2025, chen_antibody-drug_2025}. At the same time, even single-marker evaluation is demanding, as many markers are subject to multiple scoring systems, each requiring assessment of positivity in different cell populations and compartments  \citep{moehler_concordance_2025, fehrenbacher_atezolizumab_2016, garon_pembrolizumab_2015}. This makes manual quantitative evaluation increasingly laborious and prone to interobserver variability \citep{attwood_trends_2021, chang_interobserver_2019, butter_impact_2022, van_bockstal_evaluation_2024, bontoux_reproducibility_2024}. Spatial relationships such as cell–cell distances and cellular neighborhoods are also largely inaccessible to routine visual evaluation but may provide an additional layer of biological characterization \citep{schallenberg_ai-powered_2025, parra_immune_2023}.

Artificial intelligence (AI) can address these challenges, as pattern recognition algorithms can quantify even fine-grained features at scale. Recent progress has been driven by vision foundation models trained on large, heterogeneous histopathology datasets spanning diverse tissue types and staining modalities \citep{vorontsov_foundation_2024,xu_whole-slide_2024,chen_towards_2024,dippel_rudolfv_2024,alber_atlas_2025}. Fine-tuned for specific applications, these models have been shown to achieve competitive performance on selected pathology tasks \citep{campanella_real-world_2025,standvoss_atlas_2026}. Yet two limitations constrain their clinical impact. First, many downstream applications do not reach pathologist-level performance, and the few that approach it have rarely been prospectively clinically validated. Second, they are built for isolated tasks and do not integrate multimodal information across the multi-step diagnostic workflow. Agentic systems have the potential to address the latter, but their use in pathology remains early: current systems cover only parts of the diagnostic workflow, rely on vision-language models for qualitative image interpretation rather than validated quantitative measurements, and lack clinical validation \citep{ferber_development_2025, ferber_gpt-4_2024, goodell_large_2025, weishaupt_evidence-based_2026}.

To address these limitations, we developed LUCAID (Lung Cancer Agent for Integrative Diagnostics), an agent-driven multimodal AI system for lung cancer pathology that couples interactive diagnostic reasoning with a suite of extensively validated quantitative analysis modules. LUCAID's modules are based on the Atlas family of histopathology models and here we extend them towards comprehensive lung cancer precision pathology \citep{standvoss_atlas_2026,alber_atlas_2025, alber_atlas_2026}. They cover tissue quality control, tumor detection and subtyping, TME characterization, tumor cellularity assessment, immunohistochemical (IHC) phenotyping, and PD-L1 and ADC-target expression scoring. For each case, the agent selects and orchestrates the relevant diagnostic modules, interprets their outputs, and integrates them into a structured pathology report, with all quantitative measurements generated by the underlying analysis modules. Individual components are extensively evaluated against expert annotations, while the integrated system is further assessed prospectively in a real-world clinical setting.  Importantly, this validated modularity differentiates LUCAID from current vision-language models approaches that remain unreliable on fine-grained perception and counting task that quantitative pathology depends on \citep{fu_hidden_2025,yu_benchmarking_2026,kukuljan_illusion_2026,chen_pathview-bench_2026}. Figure \ref{fig1:workflow} provides an overview of LUCAID. Designed to operate in all major histological subtypes of lung cancer, LUCAID brings reproducible, quantitative AI to the full diagnostic workflow.

Using 1,620 lung cancer cases from the Institutes of Pathology at Charité – University Medical Center Berlin, University Hospital Cologne, and Ludwig-Maximilians-University Munich, together with an independent prospective multicentric validation cohort from the National Network Genomic Medicine Lung Cancer (nNGM), we show that LUCAID's modules achieve expert-level performance against large-scale ground-truth annotations. In prospective validation across clinically actionable diagnostic tasks, LUCAID achieved 93.0\% concordance with an expert-panel adjudicated reference standard, compared with 68.3–81.1\% for individual pathologists. Building on this foundation, we demonstrate how LUCAID integrates these modules into end-to-end case analysis, pointing toward a new generation of integrated computational pathology systems.

\begin{figure}[!htb]
    \centering
    \includegraphics[width=\linewidth,height=0.7\textheight,keepaspectratio,trim={0 1.5cm 0 0},clip]{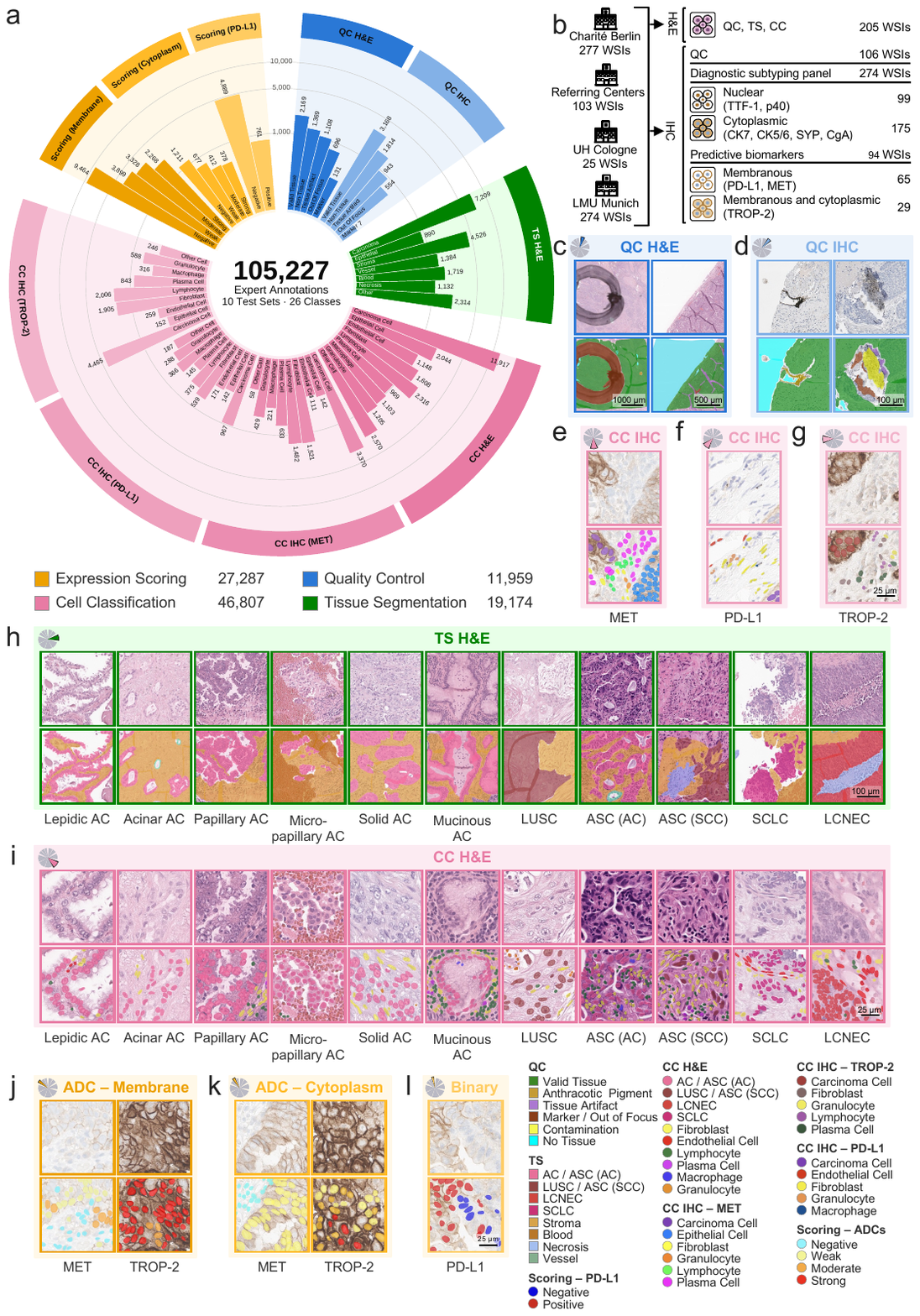}
    \small
\caption{\textbf{Expert annotation test sets for LUCAID module validation.}\newline
\textbf{a},~Distribution of 105{,}227 expert annotations across ten
independent hold-out test sets, comprising 26 prediction classes in four
algorithm categories: quality control, tissue segmentation, cell
classification and expression scoring. Annotation counts are shown for
each prediction class. \textbf{b},~Overview of the multicentric
whole-slide image test sets, showing contributing institutions and
their use for H\&E-based analysis, diagnostic subtyping and predictive
biomarker evaluation. \textbf{c--l},~Histopathological images are shown
above with the corresponding expert annotations below.
\textbf{c},\textbf{d},~H\&E and IHC examples for quality control.
\textbf{e--g},~IHC cell-classification examples for MET, PD-L1 and
TROP-2, respectively. \textbf{h},~H\&E tissue-segmentation examples
across major lung carcinoma subtypes and six adenocarcinoma subtypes.
\textbf{i},~Corresponding H\&E cell-classification examples.
\textbf{j},\textbf{k},~Subcellular expression-scoring examples for
membranous and cytoplasmic MET and TROP-2 expression. \textbf{l},~Binary PD-L1 expression-scoring
example.\newline
\textit{Abbreviations:} AC, adenocarcinoma; ADC, antibody--drug conjugate;
ASC, adenosquamous carcinoma; CC, cell classification; CgA,
chromogranin~A; H\&E, hematoxylin and eosin; IHC, immunohistochemistry;
LCNEC, large cell neuroendocrine carcinoma; LMU, Ludwig-Maximilians-University; LUSC, lung squamous cell carcinoma; MET, hepatocyte growth
factor receptor; PD-L1, programmed death-ligand 1; QC, quality control;
SCC, squamous cell carcinoma; SCLC, small cell lung cancer; SYP,
synaptophysin; TS, tissue segmentation; TROP-2, trophoblast cell-surface
antigen 2; TTF-1, thyroid transcription factor 1; WSI, whole-slide image.}
\label{fig2:annotations}
\end{figure}

\section{Results}
\label{sec:results}
LUCAID supports routine pathology workflows in lung cancer diagnosis by agentic orchestration of nine AI modules for comprehensive analysis of hematoxylin \& esoin (H\&E)- and IHC-stained WSIs: (1) \textit{tissue quality control} to identify analyzable tissue and to discard artifacts on both H\&E- and IHC-stained whole-slide images (WSIs); (2) \textit{tissue segmentation} for detecting tumor regions and delineation of tissue compartments; (3) \textit{tumor subtyping} from expressions of specific IHC markers; (4) \textit{TME profiling} by classifying individual cells on H\&E to characterize the tumor microenvironment; (5) \textit{tumor cellularity assessment} for tumor content quantification to guide molecular testing; (6) \textit{IHC cell phenotyping} by classifying cell identities IHC-stained WSIs; (7, 8) \textit{biomarker expression scoring} to quantify the expression of the prognostic biomarker PD-L1 (7) and the ADC targets MET and TROP-2 (8) at cell and subcellular resolution; and (9) \textit{structured report generation}, which integrates the quantitative results of the applicable diagnostic modules into a case-level pathology report. Each module can be called on demand during interactive pathologist–LUCAID interaction or orchestrated end-to-end to cover the diagnostic workflow from H\&E-based tissue profiling to immunohistochemical assessment and molecular testing.

To extensively assess LUCAID's capabilities, we validated it at four different levels (Figure~\ref{fig1:workflow}). First, each of the eight analysis modules was evaluated on dedicated hold-out test sets against expert ground-truth annotations. The underlying multicentric dataset comprised 1,620 lung cancer cases from the Institutes of Pathology at Charit\'e – University Medical Center Berlin, University Hospital Cologne, and Ludwig-Maximilians-University Munich and included H\&E-stained slides as well as nine different IHC markers; task-specific subsets were used for module-level testing. The dataset includes lung adenocarcinomas (LUADs), lung squamous cell carcinomas (LUSCs), adenosquamous carcinomas (ASCs), large cell neuroendocrine carcinomas (LCNECs), and small cell lung carcinomas (SCLCs). Second, to assess biological concordance, outputs of selected modules were benchmarked against orthogonal molecular reference measurements. Third, we evaluated the system's text- and report-generation capabilities by assessing whether the generated reports remain grounded in the validated module measurements. Fourth, we prospectively evaluated end-to-end execution of LUCAID in an independent lung cancer cohort from the National Network Genomic Medicine (nNGM) and benchmarked its clinically actionable diagnostic predictions against five experienced thoracic pathologists.

\subsection{Module Predictions Recover Expert Ground-Truth Annotations} 

The modules are introduced and validated in sequence of a routine diagnostic process. For tumor detection and cell phenotyping, we characterize performance on lung in detail, resolved across all five major subtypes -- LUAD, LUSC, ASC, LCNEC, and SCLC -- spanning the non-small-cell, neuroendocrine, and small-cell entities. Figure~\ref{fig2:annotations} summarizes annotation counts across the independent hold-out test sets, the multicentric WSI cohorts contributing to each diagnostic task, and representative expert annotations across quality control, tissue segmentation, cell classification and expression scoring.

\subsubsection{Module 1: Quality Control}

The quality control (QC) module automatically identifies tissue regions and common artifacts in both H\&E- and IHC-stained WSIs, distinguishing valid tissue, out-of-focus regions, tissue artifacts, markers and non-tissue background. Representative applications illustrate robust artifact detection while preserving diagnostically relevant tissue regions (Figure~\ref{fig3:qc_tumor_detection}a,b). We evaluated it on a dataset comprising 5,473 annotations from 132 H\&E WSIs and 6,486 annotations from 106 IHC WSIs. Classification performance was high across QC classes, with overall F1 scores of 0.91 for H\&E and 0.88 for IHC slides; valid tissue and non-tissue background each reached F1 scores of 0.98--1.00 in both staining modalities  (Figure~\ref{fig3:qc_tumor_detection}c). By maintaining robust performance across both staining modalities, LUCAID extends QC beyond previous H\&E-focused tools \citep{weng_grandqc_2024,jabar_fully_nodate,patil_efficient_2023,kanwal_equipping_2024}, and provides a unified preprocessing step for all downstream analyses.

\subsubsection{Module 2: Tissue Segmentation for Tumor Detection and Compartmentalization}
The tissue segmentation module automatically detects tumor regions, delineates distinct tumor compartments, and classifies different tissue types within H\&E-stained lung tissue (Figure~\ref{fig3:qc_tumor_detection}d; metastatic samples shown in Supplementary Figure~\ref{sfig2:metastasis}a), across all major lung cancer subtypes (LUAD, LUSC, ASC, SCLC, and LCNEC). It distinguishes seven categories: carcinoma, epithelium, stroma, necrosis, blood, vessel, and other tissue components commonly found in lung specimens, including alveolar lung parenchyma, cartilage, smooth muscle, nerves, and secretions such as mucus. Quantitative evaluation on a test dataset of 19,174 annotations from 205 WSIs, curated to span all five subtypes (Figure~\ref{fig2:annotations}a+h), showed strong performance across all tissue categories. The model achieved an average F1 score of 0.93 (detailed class-wise results are provided in Figure~\ref{fig3:qc_tumor_detection}e). Notably, carcinoma detection reached an F1 score of 0.98 across all carcinoma subtypes (range 0.97--0.99) and showed a comparable F1 score of 0.99 when evaluated on metastatic samples (range 0.97--0.99; Supplementary Figure~\ref{sfig2:metastasis}b).This is important, as accurately identifying tumor regions across diverse lung cancer subtypes is a central diagnostic task.

\begin{figure}[!htb]
    \centering
    \includegraphics[width=\linewidth,height=0.7\textheight,keepaspectratio,trim={0 3.0cm 0 0},clip]{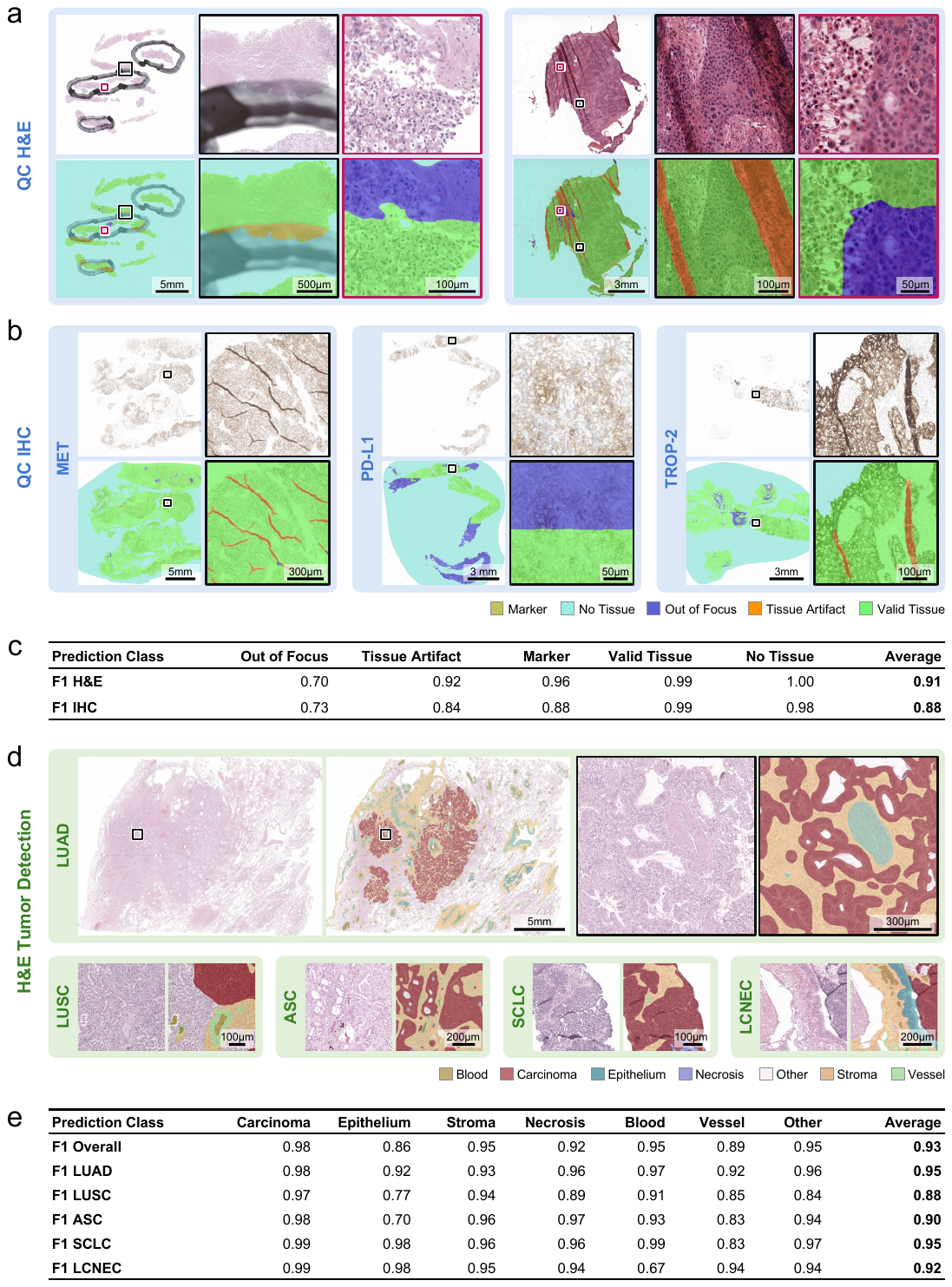}
    \small
   \caption{\textbf{Quality control and tumor detection across lung cancer subtypes.}\newline
\textbf{a},~H\&E quality-control examples shown as histopathological
images above and corresponding model-derived overlays below. Two biopsy
specimens are shown, each with an overview image followed by two
higher-magnification views. \textbf{b},~IHC quality-control examples for
MET, PD-L1 and TROP-2, each shown as an overview image and a
higher-magnification view, with histopathological images above and
corresponding model-derived overlays below. The QC model distinguishes
five classes: out of focus, tissue artifact, marker, valid tissue and no
tissue. \textbf{c},~F1 scores for each QC class on H\&E- and IHC-stained
WSIs. \textbf{d},~H\&E tumor-detection examples across five major lung
carcinoma subtypes, with histopathological images shown on the left and
corresponding model-derived tissue-segmentation overlays on the right.
For LUAD, a resection specimen is shown with a higher-magnification view;
the lower row shows LUSC, ASC, SCLC and LCNEC from left to right. The
segmentation model distinguishes seven tissue compartments: carcinoma,
epithelium, stroma, necrosis, blood, vessel and other. \textbf{e},~F1
scores for each tissue compartment overall and stratified by histological
subtype.\newline
\textit{Abbreviations:} ASC, adenosquamous carcinoma; H\&E, hematoxylin
and eosin; IHC, immunohistochemistry; LCNEC, large cell neuroendocrine
carcinoma; LUAD, lung adenocarcinoma; LUSC, lung squamous cell carcinoma;
MET, hepatocyte growth factor receptor; PD-L1, programmed death-ligand 1;
SCLC, small cell lung cancer; TROP-2, trophoblast cell-surface antigen 2.}
\label{fig3:qc_tumor_detection}
\end{figure}

\subsubsection{Module 3: Tumor Subtyping via Cell Phenotyping}
\label{sec::subtyping}

To enable cell-level analysis on IHC-stained slides, we first trained a cell phenotyping model capable of detecting and classifying individual cells across nine prediction classes in IHC WSIs (see LUCAID module 6: cell phenotyping on IHC for further details (Figure~\ref{fig4:cell_phenotyping})). Subsequently, an expression scoring model was trained to determine marker-specific expression status (positive vs. negative) exclusively in cells classified as tumor cells by the IHC cell phenotyping model. Cytoplasmic expression was evaluated for CK5/6, CK7, CgA, and SYP, whereas nuclear expression was assessed for TTF-1 and p40. Cases with >10\% marker-positive tumor cells were classified as positive for the respective marker in accordance with WHO classification criteria for lung tumors (Figure~\ref{fig5:subtyping_scoring_cellularity}a)\citep{who_classification_of_tumours_editorial_board_thoracic_2021}. Exemplary cases are shown in Supplementary Figure~\ref{sfig1:ihc_typing}. Integration of tumor cell-specific expression profiles enabled automated tumor classification according to established diagnostic criteria, such as LUAD diagnosis based on TTF-1 and CK7 positivity combined with absence of p40 and CK5/6. For lineage markers distinguishing LUAD from LUSC, LUCAID achieved accuracies of 93.3\% for TTF-1 and 97.6\% for CK7, as well as 98.1\% for p40 and 100\% for CK5/6 using pathologists' case level scoring as ground truth (Figure~\ref{fig5:subtyping_scoring_cellularity}b). For neuroendocrine markers, accuracies reached 95.5\% for SYP and 92.1\% for CgA, resulting in consistently high classification performance across all diagnostic lineage markers (Figure~\ref{fig5:subtyping_scoring_cellularity}b).

\subsubsection{Module 4: Cell Phenotyping for TME Profiling}
To enable detailed analysis of the cellular TME, the H\&E cell phenotyping module automatically detects and classifies individual cells within H\&E-stained WSIs. The model distinguishes carcinoma cells from major inflammatory cell populations, including lymphocytes, macrophages, granulocytes, and plasma cells, as well as other non-neoplastic cell types such as epithelial cells, endothelial cells, and fibroblasts. Quantitative evaluation was carried out on a test dataset of 24,880 annotations across 205 WSIs, curated to span all five major lung cancer subtypes (LUAD, LUSC, ASC, LCNEC, and SCLC; Figure~\ref{fig2:annotations}a+i). Representative cell classifications across all five major lung cancer subtypes are shown in Figure~\ref{fig4:cell_phenotyping}a, with metastatic samples provided in Supplementary Figure~\ref{sfig2:metastasis}c. Quantitative evaluation demonstrated strong performance across all cell classes, with an average F1 score of 0.95 (range across subtype: 0.90--0.98) and average class-specific F1-scores ranging from 0.89 to 0.97 (Figure~\ref{fig4:cell_phenotyping}b). Notably, carcinoma cell identification achieved an F1 score of 0.95 in primary tumor and an F1 score of 0.99 in metastatic tumors (Supplementary Figure~\ref{sfig2:metastasis}d), providing the basis for precise tumor cellularity assessment to guide molecular profiling (see section: LUCAID Module 5: Tumor Cellularity Assessment for Molecular Profiling).

\subsubsection{Module 5: Tumor Cellularity Assessment for Molecular Profiling}
The LUCAID tumor cellularity assessment module quantifies quantifies tumor content in H\&E-stained WSIs to guide molecular testing, using both cell count-based and nuclear area-based approaches.

To further capture spatial heterogeneity in tumor cellularity, we developed a tumor cell clustering-based visualization framework that generates spatial maps of local tumor content across WSIs. The resulting heatmaps provide an intuitive representation of regional variation in tumor cellularity, with a continuous color gradient ranging from low (yellow) to high (red) tumor content (Figure~\ref{fig5:subtyping_scoring_cellularity}e and Supplementary Figure \ref{sfig3:cellularity_molecular}c). The representative tumor illustrates pronounced intratumoral heterogeneity that is readily apparent in the generated heatmap but may be difficult to appreciate by conventional H\&E assessment alone. By visualizing regional variation in predicted tumor content, the heatmaps may support more standardized selection of tissue areas for molecular profiling and reduce variability associated with visual estimation.

\subsubsection{Module 6: IHC Cell Phenotyping for Biomarker Expression Scoring}
\begin{figure}[p]
    \centering
    \includegraphics[width=\linewidth,height=0.6\textheight,keepaspectratio,trim={0 8.5cm 0 0},clip]{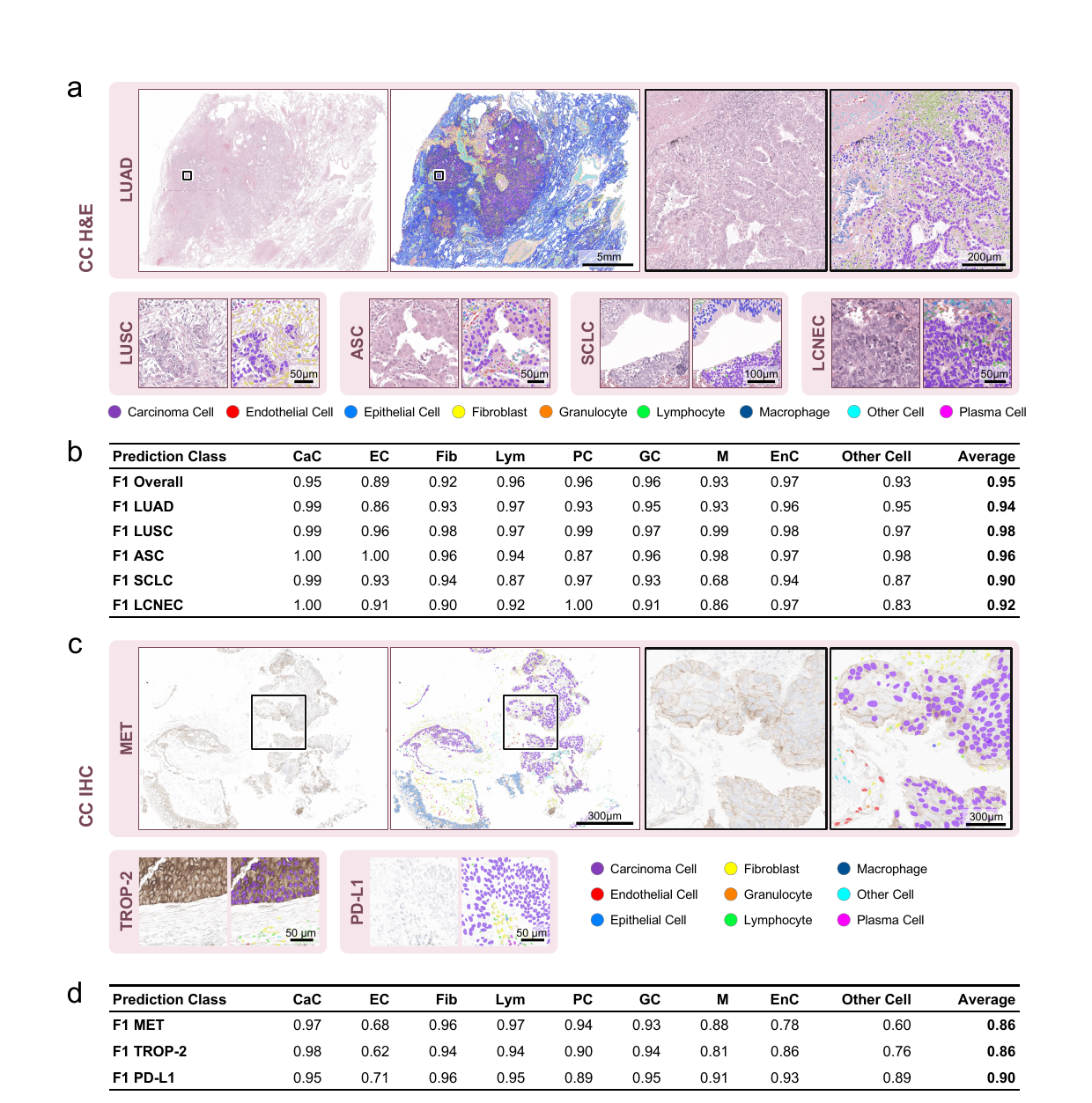}
    \small
\caption{\textbf{H\&E and IHC cell phenotyping across lung cancer subtypes and biomarkers.}\newline
\textbf{a},~H\&E cell-phenotyping examples across five major lung
carcinoma subtypes, with histopathological images shown on the left and
corresponding model-derived cell-classification overlays on the right.
For LUAD, a resection specimen is shown with a higher-magnification view;
the lower row shows LUSC, ASC, SCLC and LCNEC from left to right. The
model distinguishes nine cell classes: carcinoma cells, endothelial
cells, epithelial cells, fibroblasts, granulocytes, lymphocytes,
macrophages, plasma cells and other cells. \textbf{b},~F1 scores for each
cell class overall and stratified by histological subtype.
\textbf{c},~IHC cell-phenotyping examples for MET, TROP-2 and PD-L1 using
the same nine cell classes, with histopathological images shown on the
left and corresponding model-derived cell-classification overlays on the
right. For MET, a biopsy is shown with a higher-magnification view;
TROP-2 and PD-L1 are each shown as a single image--overlay pair.
\textbf{d},~F1 scores for each cell class stratified by
biomarker.\newline
\textit{Abbreviations:} ASC, adenosquamous carcinoma; CaC, carcinoma
cell; CC, cell classification; EC, endothelial cell; EnC, epithelial
cell; Fib, fibroblast; GC, granulocyte; H\&E, hematoxylin and eosin; IHC,
immunohistochemistry; LCNEC, large cell neuroendocrine carcinoma; LUAD,
lung adenocarcinoma; LUSC, lung squamous cell carcinoma; Lym, lymphocyte;
M, macrophage; MET, hepatocyte growth factor receptor; PC, plasma cell;
PD-L1, programmed death-ligand 1; SCLC, small cell lung cancer; TROP-2,
trophoblast cell-surface antigen 2.}
\label{fig4:cell_phenotyping}
\end{figure}

The LUCAID IHC cell phenotyping module extends the H\&E-based cell phenotyping of module 4 to IHC-stained WSIs, classifying individual cells into nine phenotypes. This allows biomarker expression to be assigned to specific cell populations rather than evaluated at the tissue level alone. Representative examples of expert test annotations are provided in Figure~\ref{fig2:annotations}e--g. We evaluated the module on the three major therapeutic biomarkers in lung cancer: PD-L1 and the ADC targets TROP-2 and MET (n = 32 TROP-2; n = 35 PD-L1; n = 30 MET). The test set comprised 3,180 ground-truth annotations for PD-L1, 10,780 for TROP-2, and 7,967 for MET (Figure~\ref{fig2:annotations}a). Annotation density reflected the increasing complexity of the respective scoring systems, from two categories for PD-L1 to four intensity levels for MET and four levels across both membranous and cytoplasmic compartments for TROP-2, with additional annotations used to capture the resulting variation in staining and expression patterns in the hold-out sets. Representative classifications across the three stainings illustrate robust cell detection and phenotyping despite differences in tumor architecture, staining intensity, and subcellular expression patterns (Figure~\ref{fig4:cell_phenotyping}c). The model achieved F1 scores of 0.86 for TROP-2, 0.86 for MET, and 0.90 for PD-L1 across all cells (Figure~\ref{fig4:cell_phenotyping}d). Notably, carcinoma-cell identification reached F1 scores of 0.98 for TROP-2, 0.97 for MET, and 0.95 for PD-L1, supporting reliable assignment of biomarker expression to the tumor-cell compartment in which these markers are clinically assessed. Together, these results support cell-level biomarker quantification across distinct therapeutic IHC stainings.

\subsubsection{Modules 7 and 8: Biomarker Expression Scoring}
\begin{figure}[p]
    \centering
    \includegraphics[width=\linewidth,height=0.7\textheight,keepaspectratio,trim={0 3.0cm 0 0},clip]{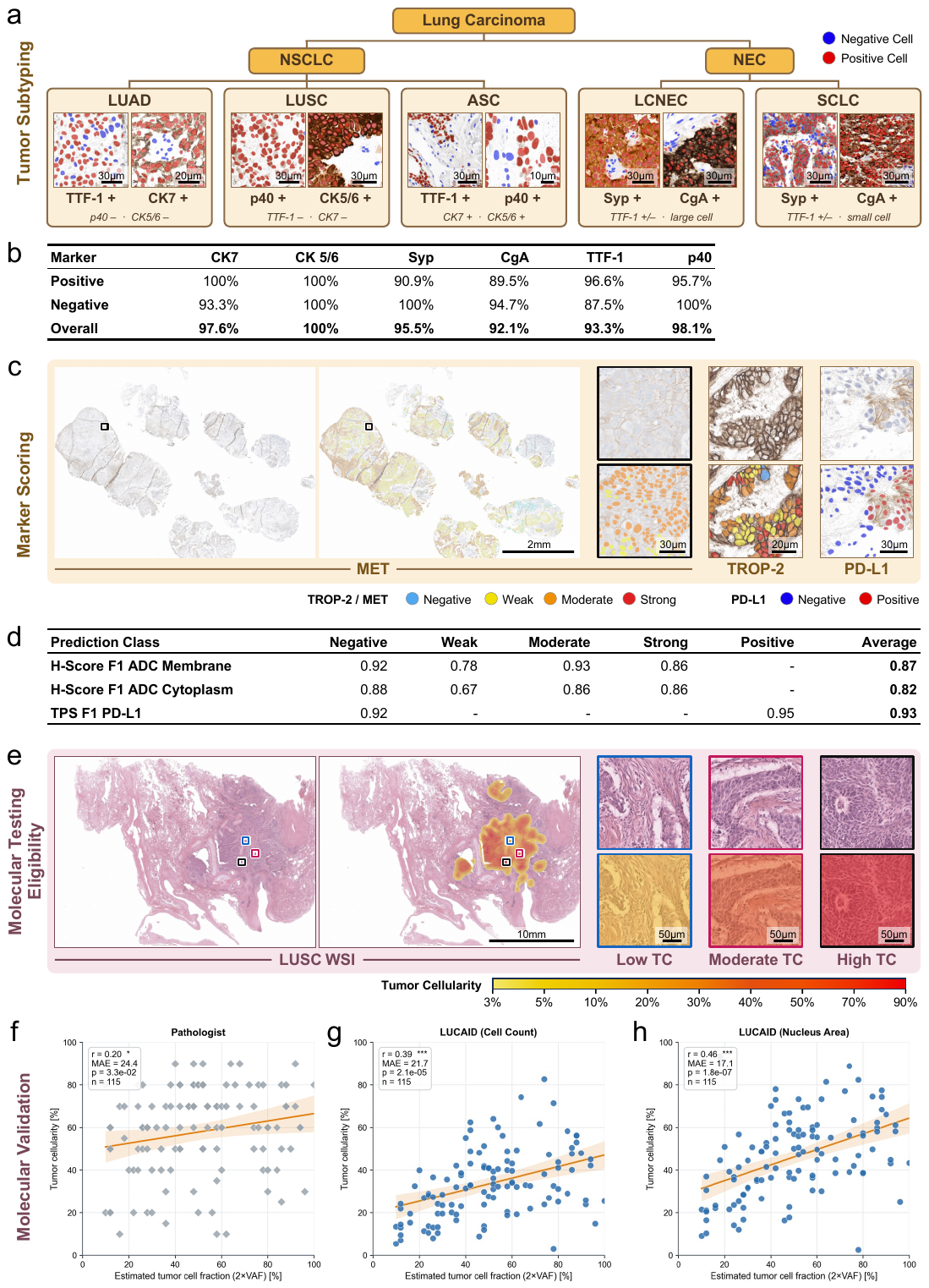}
    \small
    \caption{\textbf{Tumor subtyping, biomarker scoring and tumor cellularity for molecular profiling.}\newline
\textbf{a},~IHC-based tumor subtyping across five major lung carcinoma
entities using six diagnostic markers (CK7, CK5/6, synaptophysin,
chromogranin~A, TTF-1 and p40). For each entity, two key diagnostic
stains are shown as IHC images with model-derived cell-classification
overlays; marker-positive cells are shown in red and marker-negative
cells in blue. \textbf{b},~Accuracy for positive, negative and overall
classification for each diagnostic marker. \textbf{c},~Biomarker
expression-scoring examples for MET, TROP-2 and PD-L1. For MET, a biopsy
overview is shown with the model-derived overlay, followed by a
higher-magnification image--overlay pair; TROP-2 and PD-L1 are each shown
as paired histopathological images and model-derived overlays. MET and
TROP-2 expression is classified as negative, weak, moderate or strong,
whereas PD-L1 expression is classified as negative or positive.
\textbf{d},~F1 scores for each expression class for membranous and
cytoplasmic ADC-target scoring and PD-L1 tumor proportion scoring.
\textbf{e},~Tumor cellularity assessment for molecular testing
eligibility. A resection specimen is shown as an H\&E whole-slide
overview with the corresponding model-derived cellularity map, together
with higher-magnification examples of regions with low, moderate and high
tumor cellularity; histopathological images are shown above and
corresponding overlays below. \textbf{f--h},~Orthogonal molecular
validation of tumor cellularity estimates against \textit{KRAS} variant
allele frequency (n~=~115). Correlation between \textit{KRAS} variant
allele frequency and tumor cellularity for \textbf{f},~routine
pathologist estimates, \textbf{g},~AI-predicted tumor cell proportion and
\textbf{h},~AI-predicted tumor nuclear area proportion. Pearson $r$ and
Spearman $\rho$ (both two-tailed) and the MAE versus variant allele
frequency are shown in each panel. Orange lines indicate linear fits;
shaded bands indicate 95\% confidence intervals.\newline
\textit{Abbreviations:} ADC, antibody--drug conjugate; ASC, adenosquamous
carcinoma; CgA, chromogranin~A; CK5/6, cytokeratin 5/6; CK7, cytokeratin
7; H\&E, hematoxylin and eosin; IHC, immunohistochemistry; LCNEC, large
cell neuroendocrine carcinoma; LUAD, lung adenocarcinoma; LUSC, lung
squamous cell carcinoma; MAE, mean absolute error; MET, hepatocyte growth
factor receptor; PD-L1, programmed death-ligand 1; SCLC, small cell lung
cancer; SYP, synaptophysin; TC, tumor cellularity; TTF-1, thyroid transcription factor 1;
TROP-2, trophoblast cell-surface antigen 2.}
\label{fig5:subtyping_scoring_cellularity}
\end{figure}

Building on the LUCAID IHC cell phenotyping module, we developed expression scoring modules to quantify biomarker expression intensity at individual cell level and subcellular compartments. Intensity was classified into four categories (negative, weak, moderate, and strong) for membranous MET and TROP-2 as well as cytoplasmic TROP-2 expression. PD-L1 was assessed using a binary classification scheme (positive vs. negative). Figure~\ref{fig2:annotations}j--l presents examples of expert test annotations across the different expression scoring categories. For ADC (TROP-2 and MET) assessment, the test datasets included 18,959 (TROP-2 = 13,062 and MET = 5,897) and 2,678 annotations for membrane and cytoplasm, respectively (Figure~\ref{fig2:annotations}a). For PD-L1, 5,650 annotations were collected for testing (Figure~\ref{fig2:annotations}a). Module performance was highest for PD-L1, reaching an average F1 score of 0.93 (Figure~\ref{fig5:subtyping_scoring_cellularity}d). For MET and TROP-2, F1 scores reached 0.87 for membranous and 0.82 for cytoplasmic expression across all intensity categories. Notably, classification performance for the clinically most relevant moderate and strong expression categories ranged between 0.86 and 0.93, supporting accurate identification of cases potentially eligible for targeted therapies.
Overall, the consistent performance observed across distinct biomarkers and subcellular expression patterns highlights the strong generalizability of LUCAID expression scoring. This broad applicability is particularly relevant for emerging therapeutic targets, including the rapidly expanding class of ADC targets, where the same analytical framework can be applied across biomarkers with different staining patterns and scoring requirements.

\clearpage

\begin{figure}[p]
    \vspace{-15mm}
    \centering
    \includegraphics[width=\linewidth,height=0.7\textheight,keepaspectratio,trim={0 2.5cm 0 0}]{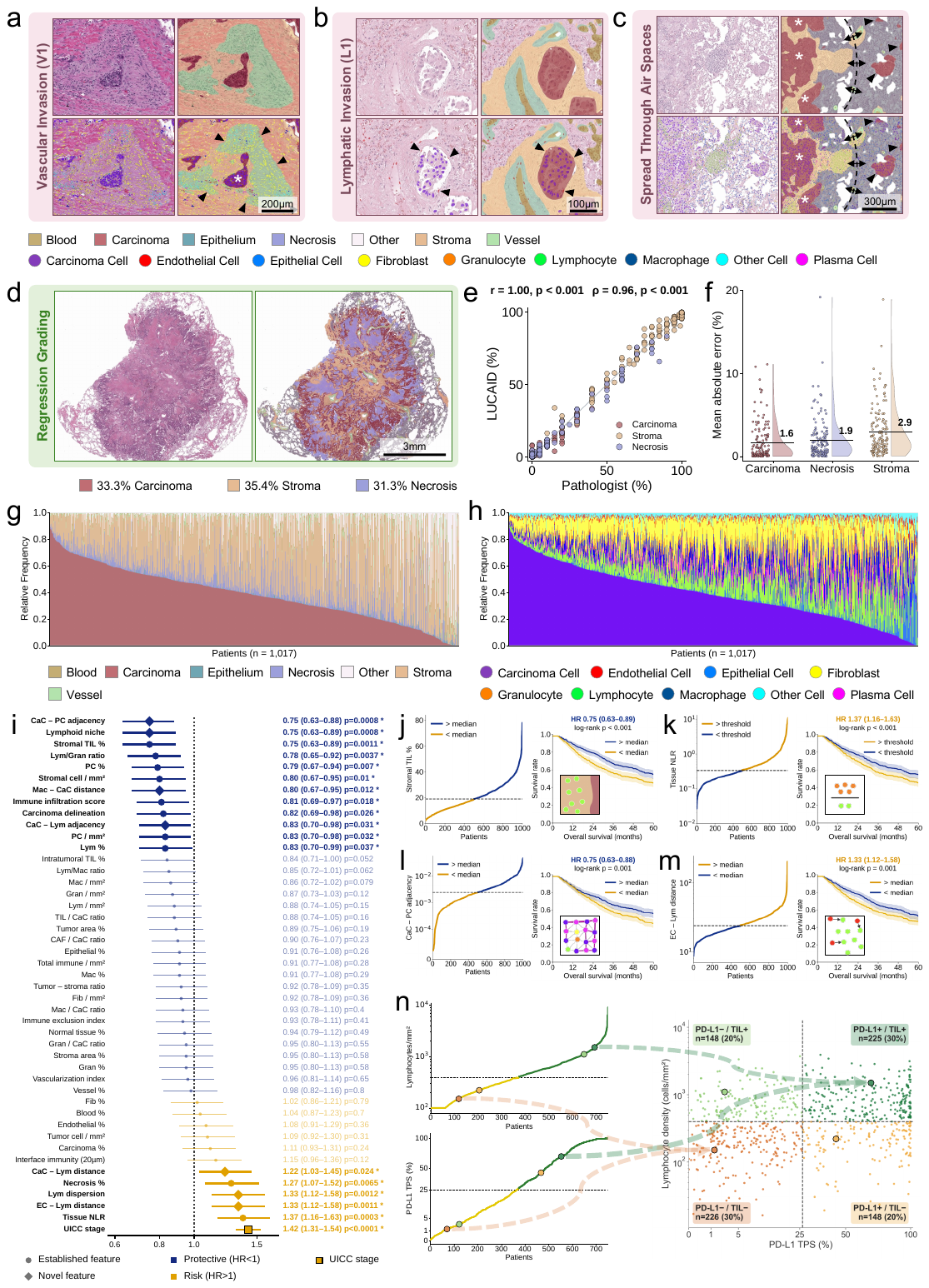}
    \small

\caption{\textbf{Clinicopathological applications and biomarker discovery in lung cancer.}\newline
\textbf{a--c},~Histopathological examples of vascular invasion
(\textbf{a}), lymphatic invasion (\textbf{b}) and spread through air
spaces (STAS; \textbf{c}), visualized using tissue-segmentation and
cell-classification overlays. For each finding, H\&E is shown at the top
left, tissue segmentation at the top right, cell classification at the
bottom left and the combined overlay at the bottom right. In vascular
invasion, black arrows indicate the detected vessel and the white
asterisk marks intravascular carcinoma. In lymphatic invasion, black
arrows indicate endothelial cells surrounding the carcinoma cells. In
STAS, white asterisks mark the main tumor, the dashed double-headed line
indicates a defined distance from the tumor border within healthy lung
parenchyma (gray), and black arrows indicate carcinoma within alveolar
spaces beyond this zone. \textbf{d},~Pathological regression grading in a
resection specimen, showing H\&E (left) and the tissue-segmentation
overlay (right) with quantification of viable carcinoma, stroma and
necrosis. \textbf{e},~Correlation between the joint pathologist
assessment and AI-predicted proportions of carcinoma, stroma and necrosis
(n~=~140). Pearson and Spearman correlation coefficients are shown; tests
are two-sided. \textbf{f},~Mean absolute error (MAE) of AI-predicted
carcinoma, stroma and necrosis proportions relative to the joint
pathologist assessment. \textbf{g},~Discovery-cohort (n~=~1{,}017)
distribution of tissue-compartment composition across tumors, ordered by
carcinoma proportion. \textbf{h},~Corresponding distribution of cell-type
composition, ordered by carcinoma-cell proportion.
\textbf{i},~UICC-adjusted Cox proportional-hazards analysis of tumor
microenvironment (TME) features. Hazard ratios are shown as points with
95\% confidence intervals; established features are shown as circles,
novel spatial features as diamonds and UICC stage as a square. Protective
and risk-associated features are shown in blue and orange, respectively;
features significant at $p < 0.05$ (stage-adjusted Wald test, uncorrected) are shown
in bold ($q$ values and per-feature $n$ in Supplementary
Table~\ref{tab:S2_forest_results}; see Methods). \textbf{j--m},~Cohort-level distributions
and corresponding Kaplan--Meier overall-survival curves stratified into high and
low groups for \textbf{j},~stromal
tumor-infiltrating lymphocyte (TIL) percentage; \textbf{k},~tissue
neutrophil-to-lymphocyte ratio; \textbf{l},~carcinoma cell--plasma cell
adjacency; and \textbf{m},~endothelial cell--lymphocyte distance; each panel annotates the
stage-adjusted hazard ratio (95\% CI) and the unadjusted log-rank $p$ (adjusted
Wald $p$ in Supplementary Table~\ref{tab:S2_forest_results}).
\textbf{n},~Cohort-level distributions of intratumoral lymphocyte density
and PD-L1 tumor proportion score (TPS), with joint classification into
four immune phenotypes defined by high or low TIL density and PD-L1
expression, which are associated with differential response to immune
checkpoint inhibition.\newline
\textit{Abbreviations:} CaC, carcinoma cell; CAF, cancer-associated
fibroblast; E, epithelial cell; Fib, fibroblast; Gran, granulocyte; H\&E,
hematoxylin and eosin; Lym, lymphocyte; Mac, macrophage; MAE, mean
absolute error; NLR, tissue neutrophil-to-lymphocyte ratio; PC, plasma
cell; PD-L1, programmed death-ligand 1; STAS, spread through air spaces;
TIL, tumor-infiltrating lymphocyte; TME, tumor microenvironment; TPS,
tumor proportion score; UICC, Union for International Cancer Control.}
\label{fig6:clinicopathological_applications}
\end{figure}

\subsection{Cellularity Estimates Correlate with Molecular Reference Data}
\label{sec:cell_estimates_correlate_w_mol_data}

To evaluate the clinical applicability of automated tumor cellularity assessment for molecular diagnostics, we analyzed a cohort of 115 \textit{KRAS}-mutated lung cancer cases with available molecular profiling data. Tumor cellularity was quantified within pathologist-annotated tumor regions selected for molecular testing (Supplementary Figure \ref{sfig3:cellularity_molecular}a), and LUCAID-derived estimates and routine pathologist assessments were compared with \textit{KRAS} variant allele frequency (VAF) as an orthogonal molecular reference for tumor-derived DNA content (see Methods: Comparison of Tumor Cellularity Assessment Variants).
Routine pathologist assessment showed only weak correlation with the molecular reference (r = 0.20, p = 0.033), resulting in a mean absolute error (MAE) of 24.42 percentage points (ppt; Figure~\ref{fig5:subtyping_scoring_cellularity}f). Using the conventional cell count-based approach, LUCAID improved correlation with \textit{KRAS} VAF to r = 0.39 (p < 0.001) while reducing the MAE to 21.7 ppt (Figure~\ref{fig5:subtyping_scoring_cellularity}g). Notably, the nuclear area-based tumor cellularity metric further strengthened agreement with the molecular reference, increasing the correlation to r = 0.46 (p < 0.001) and reducing the MAE to 17.1 ppt, corresponding to a 30.0\% reduction relative to routine pathologist assessment and a 21.2\% reduction relative to cell count-based quantification (Figure~\ref{fig5:subtyping_scoring_cellularity}h; representative cell classification heatmaps within pathologist-annotated tumor regions are shown in Supplementary Figure~\ref{sfig3:cellularity_molecular}b). Taken together, these findings demonstrate that automated cell-level analysis can improve the assessment of tumor cellularity for molecular profiling in a setting characterized by substantial interobserver variability among pathologists \citep{smits_estimation_2014, mikubo_calculating_2020, dufraing_external_2018} and suggest that nuclear area-based quantification may provide a more accurate estimate of tumor cellularity than conventional cell count-based approaches.

\subsection{LUCAID Captures Patterns of Tumor Invasion and Identifies Prognostic TME Features} 

Beyond its core diagnostic modules, we investigated whether LUCAID-derived
tissue- and cell-level measurements could capture clinically relevant
histopathological patterns and extend routine assessment toward comprehensive characterization of the TME at tissue, cellular and spatial levels.
By integrating tissue segmentation with cell phenotyping, LUCAID supported automated assessment of vascular invasion (Figure~\ref{fig6:clinicopathological_applications}a), lymphatic invasion (Figure~\ref{fig6:clinicopathological_applications}b) and spread through air spaces (STAS; Figure~\ref{fig6:clinicopathological_applications}c), established prognostic features included in routine diagnostic reporting \citep{travis_international_2024, higgins_lymphovascular_2012, kessler_blood_1996}. 
Quantification of viable tumor, necrosis, and stroma further enabled automated regression grading of resected specimens following neoadjuvant therapy according to the recommendations of  the International Association for the Study of Lung Cancer (IASLC; Figure~\ref{fig6:clinicopathological_applications}d) \citep{travis_iaslc_2020}. LUCAID
quantifications closely matched the joint pathologist assessment across all tissue compartments (pooled Pearson $r$ = 0.996,
$p < 0.001$; Spearman $\rho$ = 0.96, $p < 0.001$), with mean absolute
errors of 1.6, 1.9 and 2.9 percentage points for carcinoma, necrosis
and stroma, respectively (Figure~\ref{fig6:clinicopathological_applications}e,f). Thus, a semiquantitative visual estimate of pathological treatment response could be translated into a reproducible quantitative measurement.
Applied to 1{,}001 patients of the discovery cohort with available follow-up, LUCAID quantified the distribution of tissue, cellular and spatial TME features across tumors (Figure~\ref{fig6:clinicopathological_applications}g,h), including features that are too time-consuming for routine manual assessment or not practically quantifiable by visual inspection alone. These cohort-level distributions served as reference distributions for contextualizing individual LUCAID-derived measurements and subsequent report-level interpretation. UICC-adjusted Cox
proportional-hazards analyses recapitulated established prognostic
associations, including higher stromal tumor-infiltrating lymphocyte
(TIL) levels with improved overall survival and higher tissue
neutrophil-to-lymphocyte ratio (NLR) with adverse outcome
(Figure~\ref{fig6:clinicopathological_applications}i--k)\citep{park_artificial_2022, yan_prognostic_2024, ilie_predictive_2012}. Beyond these established markers, greater carcinoma
cell--plasma cell adjacency was associated with improved survival,
whereas greater endothelial cell--lymphocyte distance was associated with
adverse outcome, revealing additional prognostic information in the spatial organization of the TME  (Figure~\ref{fig6:clinicopathological_applications}l,m).
Finally, integrating intratumoral lymphocyte density with AI-predicted
PD-L1 tumor proportion score (TPS) stratified tumors into four immune
phenotypes defined by TIL-high/low and PD-L1-high/low status, mirroring a
previously described TME classification associated with differential
response to PD-1/PD-L1 blockade (Figure~\ref{fig6:clinicopathological_applications}n) \citep{shirasawa_differential_2021}. LUCAID provides a standardized and automated implementation of this immune phenotyping strategy by integrating H\&E-derived intratumoral lymphocyte density with IHC-based PD-L1 assessment, making combined immune phenotyping scalable across larger cohorts.

Taken together, these analyses show that LUCAID spans established diagnostic pathology, quantitative treatment-response assessment and advanced TME characterization, including invasion patterns, pathological regression grading, established prognostic markers, novel spatial features associated with outcome, and combined immune phenotypes associated with differential treatment response. The resulting cohort-wide distributions provide a reference for case-level interpretation and LUCAID report generation (Figure~\ref{fig7:report_generation}) and serve as a benchmark for future studies of the lung TME.

\subsection{LUCAID Generates Structured Pathology Reports Grounded in Validated Modules}

The outputs of LUCAID modules  provide a broad set of quantitative biological measurements that require clinical interpretation and accessible presentation for routine use. We therefore evaluated whether LUCAID could integrate these validated module outputs into structured pathology reports and whether the generated content remained faithful to the underlying measurements.

For a given case, the outputs of all applicable modules are retrieved and assembled into a report organized into predefined diagnostic sections. Each section pairs the quantitative measurements of one module with a concise interpretation of their diagnostic or therapeutic relevance. A concluding case-level assessment integrates findings across modules and summarizes their potential clinical implications and treatment considerations. Quantitative values are transferred directly from the validated analysis modules, separating the generation of measurements from their subsequent LUCAID-based clinical contextualization and interpretation.

To ground these interpretations in published evidence, we implemented a PubMed retrieval and citation mechanism. Before generating an interpretation, the agent queries PubMed~\footnote{https://pubmed.ncbi.nlm.nih.gov/}  through the NCBI E-utilities interface and and retrieves potentially relevant publications. The agent may then use these records to support selected interpretive statements. Crucially, citation entries are taken directly from PubMed, and LUCAID only selects which of the retrieved records to cite. References to non-existent publications therefore cannot appear in the report by construction. A representative report is shown in Figure~\ref{fig7:report_generation}a, with the corresponding reference list provided in Appendix~\ref{app:rep_eval}. Beyond the static report, the agent supports case-specific follow-up queries, calling relevant analysis modules as required and integrating their outputs into the response (example conversation shown in Figure~\ref{fig7:report_generation}b).

To assess report fidelity and clinical validity, we adapted and combined established evaluation frameworks \citep{singhal_large_2023,yu_evaluating_2023,ostmeier_green_2024,singhal_toward_2025} and decomposed generated reports into atomic statements. Board-certified pathologists independently evaluated each statement according to its function. Statements that directly
reference a module output are graded for groundedness, defined as preservation of the direction and magnitude of the underlying measurement (grounded, partially grounded, ungrounded). Statements providing non-trivial clinical interpretations of one or more module outputs were graded for interpretation correctness. Statements carrying a citation are additionally graded for citation correctness, assessing whether the cited publication supported the associated claim (yes, partially, no). Incorrect statements were further assessed for potential harm (none, mild-to-moderate or severe) and for the likelihood of influencing a clinical decision under routine review (low, medium or high). At the report level, we evaluated clinically relevant omissions and contradictions between section-level findings and the overall assessment. Detailed evaluation criteria and instructions are provided in Appendix \ref{app:rep_eval}.

Ten reports containing outputs from all LUCAID modules were evaluated, yielding 1{,}365 atomic statements (median, 134.5 statements per report; range, 119--154). Of these, 134 (9.8\%) directly referenced a module output and 1{,}231 (90.2\%) contained a clinical interpretation based on one or more outputs (Figure~\ref{fig7:report_generation}c). All 134 referencing statements were graded as grounded, indicating that the agent reliably anchors its statements in the module outputs rather than introducing quantities of its own. Of the 1{,}231 interpretive statements, 1{,}194 (97.0\%) were judged correct and 37 (3.0\%) incorrect, distributed over nine of the ten reports. A total of 159 citations were placed, of which 131 (82.4\%) fully or partially supported the associated statement, whereas 28 (17.6\%) did not. At the report level, no clinically relevant omissions were identified across 70 section-level reviews; however, three contradictions between the overall assessment and individual report sections occurred in two reports (Figure~\ref{fig7:report_generation}c). Harm was rated for 111 errors. Of these, 96 (86.5\%) were considered to have no potential for harm and 15 (13.5\%) mild-to-moderate potential, with none rated as severe.

\begin{figure}[p]
    \vspace{-10mm}
    \centering
    \includegraphics[width=\linewidth,height=0.9\textheight,keepaspectratio,trim={0 3.5cm 0 0}]{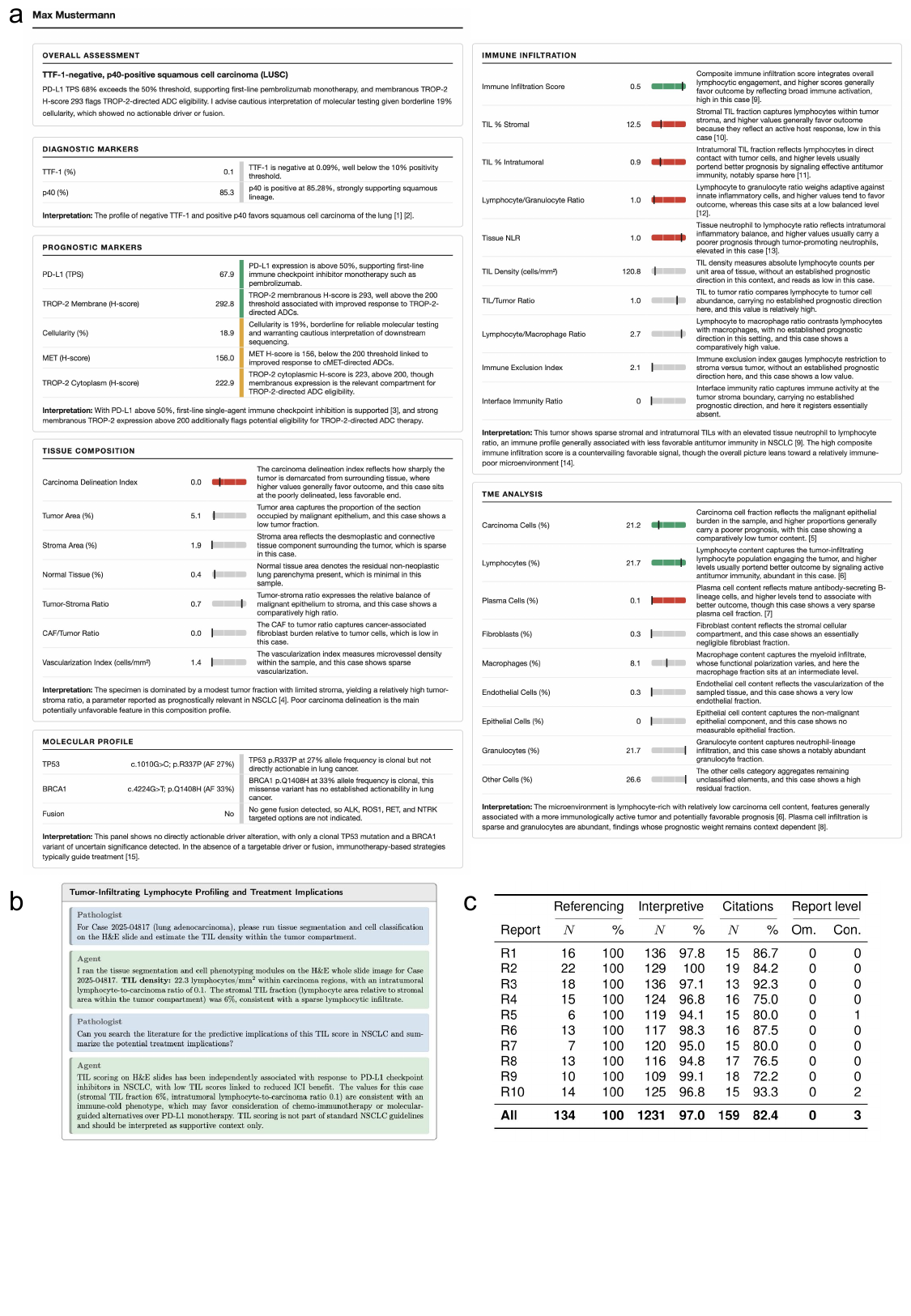}
    \small
    \caption{\textbf{Structured report generation, agentic interaction and report evaluation.}\newline
\textbf{a},~Representative structured diagnostic report generated
by LUCAID. Module outputs are integrated into predefined sections and
interpreted for clinical relevance. For tissue-composition and TME
readouts without established clinical cut-offs, values are contextualized
by their percentile position within reference distributions derived from
the discovery cohort. LUCAID integrates the section-level findings
into an overall case assessment. \textbf{b},~Representative pathologist--agent
dialogue in natural language. The pathologist requests
TIL density estimation and potential treatment implications; LUCAID provides an
evidence-grounded answer using quantitative AI module results and
retrieved literature. \textbf{c},~Evaluation of ten
generated reports at claim and report level. Reports were decomposed into
atomic statements. Referencing denotes statements that directly cite
quantitative module results; the percentage indicates the proportion
judged grounded in direction
and magnitude of the result. Interpretive denotes statements drawing a clinical
inference from one or more module results; the percentage indicates the
proportion judged correct. Citations reports the number of literature
citations and the proportion judged to (partially) support the
associated statement. At the report level, omissions (Om.) indicate
clinically relevant findings evident from the reported results but not
reflected in the summary sections. Contradictions (Con.) indicate
inconsistencies between the overall assessment and the individual report
sections.\newline
\textit{Abbreviations:} ADC, antibody--drug conjugate; Con.,
contradiction; IHC, immunohistochemical; MET, hepatocyte growth factor
receptor; NSCLC, non-small cell lung cancer; Om., omission; PD-L1,
programmed death-ligand 1; QC, quality control; TIL, tumor-infiltrating
lymphocyte; TME, tumor microenvironment; TPS, tumor proportion score;
TROP-2, trophoblast cell-surface antigen 2; TTF-1, thyroid transcription
factor 1.}
\label{fig7:report_generation}
\end{figure}

\subsection{LUCAID Shows High Concordance across Clinically Actionable Tasks}

After establishing LUCAID performance across module-level expert annotation benchmarks, orthogonal molecular validation and structured report evaluation, we next assessed its prospective clinical performance on five key tasks in lung cancer diagnostics: tumor cellularity assessment for molecular testing, PD-L1 tumor proportion scoring (TPS), MET H-score assessment, and membranous and cytoplasmic TROP-2 H-score assessment. Seventy consecutive patients with lung cancer were prospectively enrolled through the nNGM program (cohort characteristics are provided in Supplementary Table~\ref{tab:S1_cohort_characteristics}).
 
For each case, these five clinically actionable diagnostic tasks were assessed by the corresponding LUCAID modules and compared with five experienced pathologists across 328 diagnostic assessments (clinical concordance; cf. Section~\ref{sec:cac}). An expert-panel adjudicated reference standard, established through re-evaluation of all 70 cases after a washout period of at least three months, served as the reference for all comparisons (cf. Section~\ref{sec:consensus_ref}). 

\begin{figure}
    \centering
    \includegraphics[width=\linewidth,height=0.7\textheight,keepaspectratio,trim={0 4.5cm 0 0}]{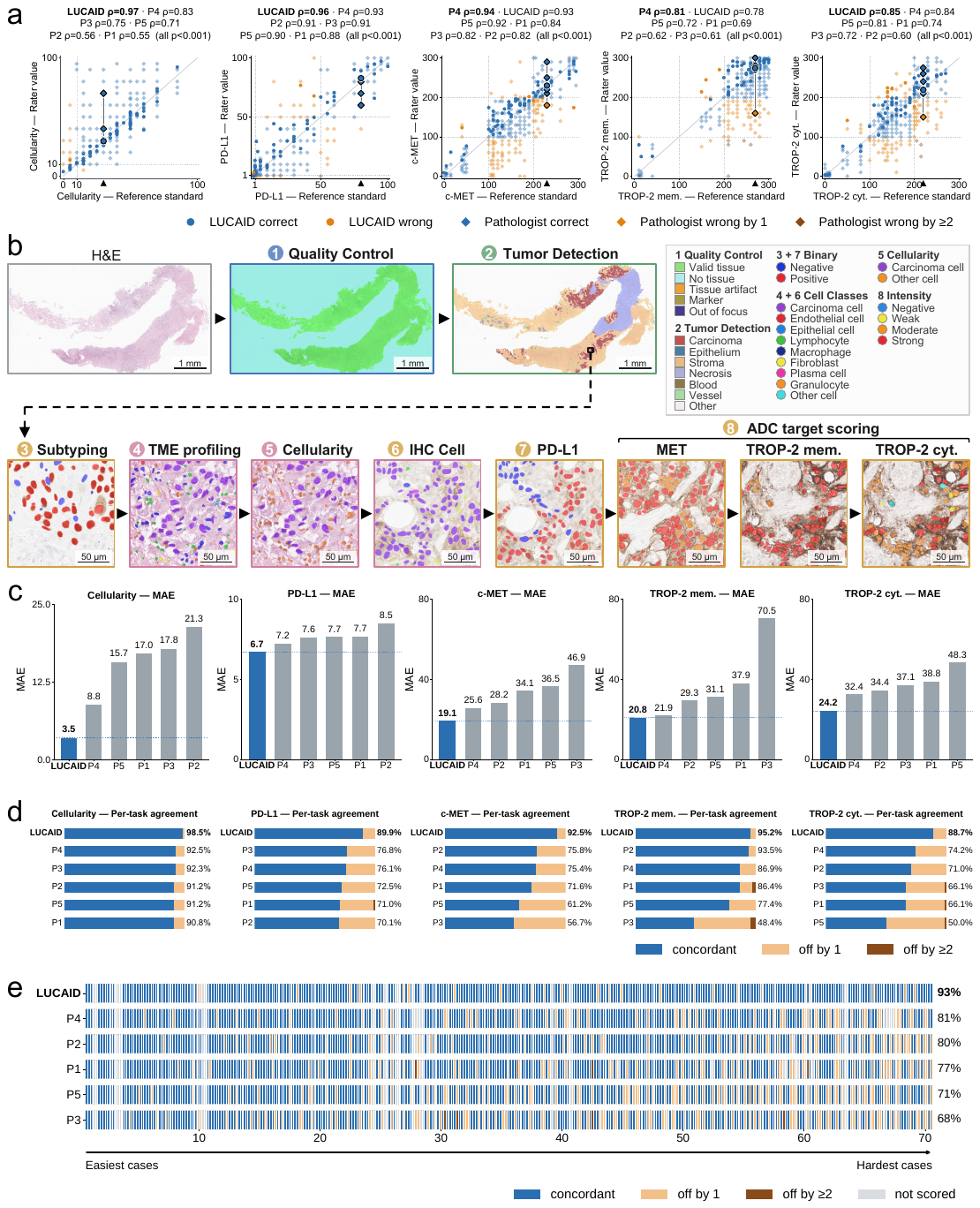}
    \small
 \caption{\textbf{Prospective clinical validation against an expert-panel adjudicated reference standard.}\newline
\textbf{a},~Rater-versus-reference comparisons for LUCAID and five
pathologists (P1--P5) across tumor cellularity, PD-L1 tumor proportion
score (TPS), MET H-score, TROP-2 membranous H-score and TROP-2
cytoplasmic H-score (numbers of evaluable cases per task in Section~\ref{sec:prosp_clin_valid_cohort}). Dashed lines indicate
clinical decision thresholds. Spearman correlation coefficients are shown
for each rater; all tests are two-sided. \textbf{b},~Representative LUAD case showing LUCAID outputs across the diagnostic
workflow, including quality control, tumor detection, tumor subtyping,
TME profiling, tumor cellularity assessment, IHC cell phenotyping, PD-L1
scoring and MET/TROP-2 scoring. Histopathological images are shown with
the corresponding model-derived overlays. \textbf{c},~Mean absolute error
(MAE) relative to the expert-panel adjudicated reference standard for
LUCAID and each pathologist across the five scoring tasks, ranked from
lowest to highest error within each task. \textbf{d},~Clinical action
concordance for LUCAID and each pathologist, defined as the proportion of
cases assigned to the same clinically relevant category as the reference
standard. Thresholds were tumor cellularity $<$10\% versus $\geq$10\%;
PD-L1 TPS $<$1\%, 1--49\% and $\geq$50\%; and MET and TROP-2 H-scores
$<$100, 100--199 and $\geq$200. Bars indicate concordant classifications
and deviations from the reference by one or at least two categories.
\textbf{e},~Case-level clinical action concordance across 70
prospectively evaluated cases, ordered from highest to lowest overall
concordance. Each case is represented by five tiles corresponding to the
five scoring tasks; colors indicate concordance, deviations from the
reference by one or at least two categories, and tasks not scored.\newline
\textit{Abbreviations:} H\&E, hematoxylin and eosin; IHC,
immunohistochemical; MAE, mean absolute error; MET, hepatocyte growth
factor receptor; PD-L1, programmed death-ligand 1; TME, tumor
microenvironment; TPS, tumor proportion score; TROP-2, trophoblast
cell-surface antigen 2.}
\label{fig8:clinical_validation}
\end{figure}

Continuous scores showed strong correlations with the reference across all five tasks, with LUCAID Spearman correlations ranging from $\rho$ = 0.78 to 0.97 and pathologist correlations from $\rho$ = 0.55 to 0.94 (all two-sided;
$\rho < 0.001$; Figure~\ref{fig8:clinical_validation}a). Across the six raters, LUCAID ranked third
overall by correlation strength, indicating agreement within the range of inter-rater variability observed across evaluators (Supplementary Figure~\ref{sfig4:corr}a). A representative lung adenocarcinoma
case illustrates the corresponding LUCAID assessments across the
diagnostic workflow (Figure~\ref{fig8:clinical_validation}b).

Differences between evaluators became more apparent when considering
quantitative error. LUCAID showed the lowest mean absolute error (MAE)
for tumor cellularity assessment (3.47 percentage points;
$\rho$ = 0.97), compared with 8.8--21.26 percentage points for the
pathologists. For the remaining tasks, LUCAID achieved MAEs of 6.70 percentage points for
PD-L1 ($\rho$ = 0.96), 19.11 for MET ($\rho$ = 0.93), 20.82 for
membranous TROP-2 ($\rho$ = 0.78) and 24.20 for cytoplasmic TROP-2
($\rho$ = 0.85), placing LUCAID among the lowest-error evaluators
across tasks (Figure~\ref{fig8:clinical_validation}a,c). For tumor cellularity, evaluation against
\textit{KRAS} variant allele frequency (VAF) as an orthogonal molecular
reference further supported the nuclear area-based approach (MAE = 9.15
percentage points, $r$ = 0.82), compared with the tumor cell count-based
approach (MAE = 15.71 percentage points, $r$ = 0.84) and pathologist
estimates (MAE = 14.33--22.11 percentage points; $r$ = 0.20--0.69;
Supplementary Figure~\ref{sfig4:corr}b).

At clinically relevant decision thresholds, LUCAID achieved an overall
clinical action concordance of 93.0\% with the expert-panel adjudicated
reference standard, compared with 68.3--81.1\% for the individual
pathologists (Figure~\ref{fig8:clinical_validation}d). Concordance was 98.5\% for molecular testing
eligibility based on tumor cellularity ([0--10\%) versus [10--100\%]),
compared with 90.8--92.5\% for the pathologists; 89.9\% for PD-L1
treatment stratification (TPS [0--1\%), [1--50\%) and [50--100\%]),
compared with 70.1--76.8\%; and 92.5\% for MET classification (H-score
[0--100), [100--200) and [200--300]), compared with 56.7--75.8\%. The
greatest interobserver variability was observed for TROP-2, with
concordance of 95.2\% for membranous expression versus 48.4--93.5\% among
pathologists and 88.7\% for cytoplasmic expression versus 50.0--74.2\% (Figure~\ref{fig8:clinical_validation}d).

At the case level, interobserver variability became even more pronounced across the prospective cohort. At least one deviation from the reference category occurred in 67 of 70 cases (96\%), whereas only three cases (4\%) showed complete concordance across all raters and evaluated tasks (Figure~\ref{fig8:clinical_validation}e). Overall, 163 of 328 assessments (50\%) showed disagreement among the five pathologists, with the extent of disagreement varying markedly across cases. Discrepant cases revealed clinically relevant threshold effects. In low-cellularity tumors, pathologists more often estimated tumor content above the 10\% threshold, whereas LUCAID remained below it (representative cases shown in Supplementary Figure~\ref{sfig5:discordance}a,b), consistent with reports of visual overestimation of tumor fractions \citep{smits_estimation_2014, mikubo_calculating_2020, dufraing_external_2018}. By avoiding this upward shift across the molecular-testing threshold, LUCAID may reduce the risk of false-negative molecular testing associated with overestimation of tumor content. For PD-L1, LUCAID identified small populations of positive tumor cells crossing the 1\% threshold that were missed visually, potentially reducing the risk of underclassifying patients for immune checkpoint inhibitor therapy (representative case shown in Supplementary Figure~\ref{sfig5:discordance}c). Similar threshold-related discrepancies were observed for MET and TROP-2 (representative cases shown in Supplementary Figure~\ref{sfig5:discordance}d--f).

In summary, prospective clinical validation in a real-world precision oncology setting showed consistently high LUCAID concordance with the expert-panel adjudicated reference standard across all five clinically actionable diagnostic tasks. By providing standardized quantitative assessment across tasks, LUCAID may reduce observer-related variability and improve consistency particularly in borderline cases near clinically relevant decision thresholds.

\section{Discussion}

Precision oncology in lung cancer increasingly depends on the integration of a growing number of complex histological, immunohistochemical and molecular biomarkers. Many of these assessments are time-consuming and prone to interobserver variability. In addition, biologically relevant quantitative and spatial information contained in routine tissue sections remains difficult or impractical to assess visually and is therefore largely unused for clinical interpretation and biomarker discovery. AI offers a means to address these challenges at scale, but the field has lacked models that reliably achieve expert-level performance and systems that combine such capabilities across the diagnostic workflow rather than addressing isolated tasks. To close this gap, we developed a suite of individual AI modules that cover each step of the diagnostic process and achieve pathologist-level accuracy across all major histological lung cancer subtypes. We then introduced an agent that orchestrates these models to form LUCAID, an agentic system that lets pathologists run complex multi-step analyses through natural language, interpret the findings in biological context, and generate structured pathology reports. In prospective, multicenter validation on a real-world lung cancer cohort, LUCAID's end-to-end execution achieved 93.0\% concordance with the expert-panel adjudicated reference standard across clinically actionable decision tasks, compared with 68.3–81.1\% for five experienced thoracic pathologists.

At the core of LUCAID lies a collection of modules validated against large-scale expert annotations that together span the diagnostic workflow, from tissue quality and tumor detection through subtyping, TME characterization, and tumor cellularity assessment to subcellular biomarker scoring. Related prior approaches such as GrandQC \citep{weng_grandqc_2024} for quality control, the nnU-Net-based segmentation models of \citet{kludt_next-generation_2024} and \citet{spronck_tissue_2025}, or HistoPLUS \citep{adjadj_towards_2025} for H\&E cell phenotyping address only isolated steps of the pathology workflow and none has been prospectively validated against pathologists on the threshold-based decisions that determine treatment. The combination of task coverage and performance that the LUCAID modules provide is unprecedented in lung cancer and we show that this unlocks clinically relevant applications that had previously remained out of reach. 

First, LUCAID provides standardized quantitative assessment of clinically actionable tasks that guide systemic therapy selection in lung cancer, including molecular testing eligibility, PD-L1 tumor proportion scoring, and ADC-target expression scoring for MET and TROP-2. This contrasts with conventional semiquantitative visual assessment, which is subject to substantial intra- and interobserver variability \citep{butter_impact_2022, bontoux_reproducibility_2024, wu_interobserver_2025}. The prospective cohort illustrates the clinical relevance of this variability: the five pathologists' clinical-action categories were non-unanimous in 163 of 328 assessments, and at least one deviation occurred in 67 of 70 cases. These findings suggest that an important contribution of AI in pathology may lie not only in automation, but also in improving standardization and reproducibility of complex diagnostic assessments across observers and institutions.

Second, because the modules generalize across histological types, quantify fundamental biological features such as cell phenotypes and tissue compartments, and readily provide measurements at cohort scale, they enable both validation and discovery of biomarkers that LUCAID has not been explicitly trained for. Importantly, this includes routine applications such as pathological regression grading after neoadjuvant therapy, which can be derived directly from quantitative tissue-compartment measurements and for which LUCAID reproduced the joint pathologist assessment of viable tumor, necrosis and stroma within 1.6–2.9 percentage points. Beyond this, quantitative analysis at this scale opens up assessments that have largely remained research constructs because they require reliable quantification of multiple tissue and cellular features, including combined immune phenotypes based on intratumoral lymphocyte density and PD-L1 expression and the discovery of potentially new spatial biomarkers such as endothelial cell–lymphocyte distance. We also show that these measurements can be placed into cohort-level reference distributions across 1,001 lung cancer patients, providing an empirical context for individual patient values.  This may help pathologists interpret measurements for which established clinical cut-offs are not available. These reference distributions are incorporated into the generated reports to contextualize individual cases. Taken together, the broad variety of applications that LUCAID supports by quantifying fundamental biological building blocks of tissue, suggests that one way forward for computational pathology could be to focus on developing models that comprehensively and reliably estimate these quantities across indications and institutions. This would stand in contrast to the task-specific approaches that have so far been dominant in the field \citep{oner_obtaining_2022, el_nahhas_regression-based_2024, patkar_predicting_2024}, and would shift the emphasis for practitioners from endpoint-specific model development towards deriving biological insight from reliable, generalizable measurements.

Third, complete workflow coverage allows the agent to assemble a structured report that integrates all relevant case parameters across the whole diagnostic workflow.
The reports LUCAID generates are grounded more rigorously than previous work has been able to achieve. Systems that generate reports from whole-slide images generally reason directly over the image with a language or vision–language model \citep{lu_multimodal_2024, weishaupt_evidence-based_2026, vorontsov_end--end_2026}, which allows flexible conversation but is ill-suited to the threshold-based quantitative decisions at the core of precision oncology. Tumor proportion scores and staining intensities in specific subcellular compartments require reliable counting and density estimation across large tissue areas, while vision–language models operate on limited fields of view and do not perform reliably on fine-grained spatial tasks \citep{fu_hidden_2025, yu_benchmarking_2026, kukuljan_illusion_2026, chen_pathview-bench_2026}. Although the need for grounding and reliability is widely recognized, existing approaches ground statements in visual or literature evidence rather than in standardized biological metrics with known performance characteristics. For example, QCAgent verifies its statements by re-retrieving the slide regions that support them, and PathPocket traces interpretations to a corpus of published evidence \citep{wang_qcagent_2026, xu_multimodal_2026}. Both cases attempt to make the evidence  retrievable, but neither reliably quantifies the underlying biology. LUCAID instead builds the report directly on validated measurements. The agent contextualizes the module outputs against established clinical thresholds and provides the diagnostic and therapeutic interpretation, while the values in the report are transferred directly from the modules that generated them. In this way, failure modes that generally make generated reports hard to trust, such as the hallucination of plausible biomarker values from a diagnostic label, are excluded by construction.

This is supported by our report evaluation, in which board-certified pathologists decomposed ten generated reports into 1,365 atomic statements. Of these, 134 (9.8\%) referenced a module output directly and all were faithful to the underlying value. Furthermore, of the 1,231 (90.2\%) statements that contained a clinical interpretation, 1,194 (97.0\%) were judged correct. Among errors rated for harm, 15 (13.5\%) carried mild-to-moderate potential and none was rated severe. 
 These findings offer initial evidence that restricting text generation to the bounds set by validated module measurements may improve the reliability of language-model-based clinical reporting. Additionally, because LUCAID’s language-based interpretation is separate from the measurements displayed alongside it, erroneous reasoning remains visible to the reviewing pathologist. Citation placement was assessed in the same way, with 131 of 159 citations (82.4\%) supporting or partially supporting the statement they accompanied. Since entries are taken from PubMed, fabricated references cannot occur by construction, while improving selection of the most relevant retrieved records and incorporating additional literature databases remain natural next steps.
 
Our approach also has limitations. It focuses on lung cancer, and extending the modules to further tumor types is a logical next step. While the system was validated on a multicentric cohort and prospectively against pathologists, its influence on real-world clinical decision-making and its use by pathologists in routine practice remain to be established. Similarly, the evaluation of the generated reports rests on ten cases and should be read as a proof of principle rather than a full-scale clinical validation. Finally, the orchestrating agent is built on a general-purpose model, and domain-specific models, fine-tuning and refinement through human feedback represent straightforward routes to further improvement.

Looking beyond these open questions, our work points toward a broader vision for how AI might be integrated into pathology, and takes a concrete step in that direction. For AI to earn trust in clinical settings, it must perform sequential analyses across the entire whole-slide image reproducible across patients and interpretable to pathologists. Routine adoption additionally depends on a natural language interface, that allows such systems to be used within pathology workflows without requiring technical expertise. LUCAID was designed to meet both requirements, pairing precise and reproducible cell-level quantitative analysis across the whole slide with a conversational interface that a pathologist can direct without technical expertise. 
In the longer term, we envision a pathology ecosystem in which independently validated tools can be flexibly combined for the needs of an individual case and orchestrated through natural language. LUCAID provides a concrete realization of this concept in lung cancer as an agentic system that integrates validated quantitative AI across the whole diagnostic workflow.

\section{Methods}

\subsection{Clinical Cohorts}

Cohorts were assembled to maximize morphological and technical heterogeneity and thereby support model generalization across tissue types, staining modalities, and imaging domains.

\subsubsection{Test Cohorts for Quality Control, Tissue Segmentation, Cell Phenotyping and Biomarker Scoring}

The H\&E tissue segmentation and cell phenotyping test cohort consisted of 158 primary lung tumor WSIs ($n = 117$ from Charit\'e -- University Medical Center Berlin and $n = 41$ from external referring centers)  and 47 metastic lung tumors ($n = 39$ from Charit\'e -- University Medical Center Berlin and $n = 8$ from external referring centers) scanned across two scanner platforms (Aperio GT 450 DX and Aperio AT2 (Leica Biosystems)).
The H\&E QC cohort comprised 132 WSI across ($n = 83$ from Charit\'e -- University Medical Center Berlin and $n = 49$ from external referring centers) across two scanner platforms (Aperio GT 450 DX and Aperio AT2 (Leica Biosystems)), while the IHC QC cohort comprised 106 WSIs ($n = 52$ from Charit\'e -- University Medical Center Berlin and $n = 54$ from external referring centers) representing a broad spectrum of tumor entities,  including  lung carcinoma ($n = 72$). WSIs were scanned across eleven scanner platforms (Aperio GT 450 DX, Aperio AT2 and Aperio ScanScope (Leica Biosystems); VENTANA DP 200, VENTANA DP 600 and VENTANA iScan (Roche Diagnostics); Pannoramic 1000, Pannoramic 250 Flash III, Pannoramic SCAN II and Pannoramic MIDI II
(3DHISTECH); and Vectra Polaris (Akoya Biosciences)).

For IHC cell phenotyping and biomarker scoring, primary lung carcinoma cases were scanned on two platforms
(Pannoramic 1000 (3DHISTECH); VENTANA DP 600 (Roche Diagnostics)). Cases were distributed across tasks as follows: TROP-2 cell phenotyping, $n = 32$ ($n = 22$ from Charit\'e -- University Medical Center Berlin and $n = 10$ from University Hospital Cologne); MET cell phenotyping, $n = 30$ ($n = 20$ from Charit\'e --
University Medical Center Berlin and $n = 10$ from University Hospital Cologne); PD-L1 cell phenotyping, $n = 35$ ($n = 30$ from Charit\'e -- University Medical Center Berlin and $n = 5$ from University Hospital Cologne); TROP-2 cytoplasmic expression scoring, $n = 29$ ($n = 19$ from Charit\'e -- University Medical Center Berlin and $n = 10$ from University Hospital Cologne); membranous expression scoring, $n = 59$
(TROP-2, $n = 29$: $n = 19$ from Charit\'e -- University Medical Center Berlin and $n = 10$ from University Hospital Cologne; MET, $n = 30$: $n = 20$ from Charit\'e -- University Medical Center Berlin and $n = 10$ from University Hospital Cologne); and PD-L1 expression scoring, $n = 35$ ($n = 30$ from Charit\'e -- University Medical Center Berlin and $n = 5$ from University Hospital Cologne). 

\subsubsection{Discovery Cohort}

The discovery cohort comprised 1{,}017 surgically resected NSCLC tumors collected between 2006 and 2019 at Charit\'e -- University Medical Center Berlin and University Hospital Cologne. Of these, 1{,}001 patients had a known survival status and an overall survival of at least 3 months and were included in the analyses, comprising 581 lung adenocarcinomas (LUAD), 402 lung squamous cell carcinomas (LUSC) and 18 adenosquamous carcinomas (ASC). Clinicopathological characteristics are summarized in Supplementary
Table~\ref{tab:supp2_cohort}.

\subsubsection{Prospective Clinical Validation Cohort}
\label{sec:prosp_clin_valid_cohort}

For prospective clinical validation, 70 consecutive lung cancer cases
enrolled in the nNGM program were prospectively analyzed between November
2024 and March 2025. Cases originated from the Institute of Pathology,
Charit\'e -- University Medical Center Berlin ($n~=~16$), the Department of
Neuropathology, Charit\'e -- University Medical Center Berlin ($n~=~2$),
and five referring pathology sites in Berlin and Brandenburg, Germany ($n~=~52$).
Clinicopathological characteristics are summarized in Supplementary
Table~\ref{tab:S1_cohort_characteristics}. Comprehensive molecular
profiling according to the nNGM diagnostic workflow was performed for all
cases (see Molecular Analysis section for further details).  The number of totally assessed cases per task was 68 for tumor cellularity, 69 for PD-L1, 67 for MET, and 62 each for membranous and cytoplasmic TROP-2. 

\subsection{Immunohistochemical Tumor Classification}

For immunohistochemical lung cancer classification, a marker panel comprising
Thyroid Transcription Factor-1 (TTF-1), p40, cytokeratin 7 (CK7), cytokeratin
5/6 (CK5/6), synaptophysin (SYP), and chromogranin A (CgA) stained at the Ludwig-Maximilians-University Munich was assembled. A
total of 274 marker-specific WSIs were available for analysis, encompassing both
positive and negative staining patterns for each marker. Marker-specific
datasets included 45 TTF-1 stained WSIs (29 positive, 16 negative), 54 p40
stained WSIs (23 positive, 31 negative), 41 CK7 stained WSIs (26 positive, 15
negative), 52 CK5/6 stained WSIs (24 positive, 28 negative), 44 SYP stained WSIs
(22 positive, 22 negative), and 38 CgA stained WSIs (19 positive, 19 negative).
Cytoplasmic and membranous staining patterns were evaluated for CK7, CK5/6,
synaptophysin, and chromogranin A, whereas nuclear expression was assessed for
TTF-1 and p40. Cases with $\geq 10\%$ positive tumor cells were classified as
positive for the respective marker. Marker combinations were integrated
according to established diagnostic criteria to assign histological subtypes,
including LUAD, LUSC, ASC, Large-cell neuroendocrine
carcinoma (LCNEC), and Small-cell lung carcinoma (SCLC). LUAD was
immunohistochemically defined by CK7 positivity, absence of p40, absent or
limited CK5/6 expression, with or without TTF-1 expression. LUSC was
characterized by p40 and CK5/6 positivity together with absent or limited CK7
staining. ASC was defined by the presence of both LUAD and LUSC components, with
each component representing $\geq 10\%$ of the tumor. LCNEC was defined by
large-cell morphology and positivity for at least one neuroendocrine marker (SYP
or CgA), while lacking p40 and showing absent or limited CK7 staining. SCLC was
defined by small-cell morphology and positivity for at least one neuroendocrine
marker, together with absence of p40 and CK5/6 and absent or limited CK7
staining \citep{rossi_ttf-1_2004,kim_best_2013,yatabe_best_2019,who_classification_of_tumours_editorial_board_thoracic_2021}.

\subsection{Comparison of Tumor Cellularity Assessment Variants}

In routine molecular pathology workflows, tumor-rich tissue regions are selected and annotated on tissue slides prior to sequencing, and tumor cellularity is estimated by pathologists as a surrogate for the fraction of tumor-derived DNA within the analyzed sample. Conventionally, tumor cellularity is determined as the proportion of tumor cells among all nucleated cells within the selected region. However, this approach assumes that tumor and non-neoplastic cells contribute equally to the overall DNA content independent of differences in nuclear size.
We therefore evaluated an alternative nuclear area-based approach, based on the hypothesis that the relative nuclear area occupied by tumor cells may better approximate tumor-derived DNA content than cell counts alone. To compare the biological relevance of both approaches, \textit{KRAS} variant allele frequency (VAF) was used as an orthogonal molecular reference for tumor DNA content. As oncogenic \textit{KRAS} mutations are frequently clonal and typically heterozygous, the observed VAF is expected to approximate half of the tumor cell fraction under idealized conditions (for example, a \textit{KRAS} VAF of 20\% corresponding to approximately 40\% tumor cellularity), with the limitation that this relationship is affected by copy-number alterations, allelic imbalance, subclonality, tumor heterogeneity, and technical factors.
For this analysis, H\&E-stained WSIs from 115 lung cancer patients enrolled in the nNGM program were collected at Charit\'e -- University Medical Center Berlin, Germany, between January 2020 and December 2024. Routinely assessed diagnostic WSIs were used, in which tumor-rich regions had been annotated by board-certified pathologists and tumor cellularity had been estimated for each selected region prior to molecular testing.
Tumor cellularity was subsequently quantified by the AI pipeline within the same annotated tissue regions using two complementary approaches: (1) the proportion of carcinoma cells among all detected cells and (2) the proportion of carcinoma nuclear area relative to the total nuclear area of all detected cells. AI-derived estimates and routine pathologist assessments were compared with \textit{KRAS} VAF using Pearson and Spearman correlation coefficients and mean absolute error (MAE; see Statistical Analysis section for further details). Cases with an estimated tumor fraction (2 $\times$ VAF) exceeding 100\% were excluded from this comparison.

\subsection{Immunohistochemical Analysis}
Immunohistochemical staining was performed using a BenchMark XT automated immunostainer (Ventana Medical Systems). Antigen retrieval was conducted using either CC1 mild buffer (Ventana Medical Systems) or ER2 buffer (Leica Biosystems) at 100 °C for 16-64 minutes, according to antibody-specific requirements. Slides were incubated with primary antibodies at room temperature for 60 minutes: anti-chromogranin A (EP38, Epitomics, 1:100), anti-cytokeratin 5/6 (QR027\&QR028, Quartett, 1:300), anti-cytokeratin 7 (OV-TL 12/30, Dako, 1:1000), anti-MET (SP44, Roche/Ventana, ready-to-use),  anti-p40 (SP225, Roche/Ventana, ready-to-use), anti-PD-L1 (E1L3N, Cell Signaling, 1:100), anti-synaptophysin (27G12, Leica, 1:50), anti-TROP-2 (01, ENZ-ABS380, Enzo Life Sciences, 1:500), and anti-TTF1 (8G7G3/1, Zytomed, 1:100). Detection was performed using the avidin-biotin complex method with 3,3'-diaminobenzidine (DAB) as chromogen. Sections were counterstained with hematoxylin and bluing reagent (Ventana Medical Systems) for 12 minutes. 
For cell classification analyses, reference labels for model testing were generated by assessing immunohistochemical expression of CgA, CK5/6, CK7, p40, PD-L1, SYP, and TTF-1 using a binary classification (positive or negative). MET and TROP-2 expression was classified into four intensity categories: negative, weak, moderate, and strong. For diagnostic tumor subtyping based on immunohistochemical marker combinations, see the Tumor Subtyping via Cell Phenotyping section (Section~\ref{sec::subtyping} ). For prospective clinical validation, membranous and cytoplasmic TROP-2 expression as well as MET expression were independently assessed by five pathologists using the H-score scoring system. The percentage of positive tumor cells was estimated from 0 to 100\%, and staining intensity was classified as 0 (negative), 1 (weak), 2 (moderate), or 3 (strong). H-scores were calculated as: [(percentage of weakly positive tumor cells) × 1] + [(percentage of moderately positive tumor cells) × 2] + [(percentage of strongly positive tumor cells) × 3] \citep{anders_adc_2026}. PD-L1 expression was assessed using the tumor proportion score (TPS), defined as the percentage of PD-L1-positive tumor cells \citep{reck_pembrolizumab_2016, mok_pembrolizumab_2019}. 

\subsection{Image Analysis Pipeline}
All image-processing modules are built on the Atlas histopathology foundation model \citep{alber_atlas_2025, alber_atlas_2026} and follow the Atlas H\&E-TME development framework \citep{standvoss_atlas_2026}, to which we refer for the foundation model pre-training and for the architecture and training of the H\&E tissue-profiling modules (quality control, tissue segmentation, and H\&E cell phenotyping). The modules additionally developed as part of LUCAID -- IHC cell phenotyping and expression scoring for PD-L1, MET, and TROP-2 -- were trained within the same framework on dedicated pathologist annotations in IHC-stained whole-slide images; the IHC cell phenotyping module applies the same design as its H\&E counterpart. 
The pipeline operates as a sequential workflow: (1) quality control identifies and masks artifact-free tissue regions, (2) tissue segmentation operates on QC-validated regions to delineate tumor boundaries, (3) cell detection and phenotyping operates within segmented tissue regions excluding blood and necrotic areas, and (4) for IHC slides, expression scoring operates on phenotyped carcinoma cells. Each module passes spatial coordinates and binary masks to downstream modules, ensuring that subsequent analyses are restricted to relevant tissue areas.

\subsection{IHC Expression Scoring}
To assess IHC expression in PD-L1, MET, and TROP-2 stained images, cells classified as carcinoma by the phenotyping model were further classified by staining intensity using a second classification model following the same architectural design. For MET and TROP-2, cells were classified into four intensity categories (negative, weak, moderate, strong) based on the membranous staining; for TROP-2, cytoplasmic staining was additionally evaluated using the same intensity categories. For PD-L1, cells were classified into two categories (negative, positive) for membranous staining. Slide-level scores (H-scores or tumor proportion scores) were then calculated by aggregating cell-level predictions across all classified tumor cells.

\subsection{Agentic Orchestration and Report Generation}

Each diagnostic module was exposed as an independently callable tool for LLM-based agent models via the Model Context Protocol (MCP), together forming the LUCAID toolbox. Tool orchestration was performed by Claude Opus 4.8. 

For a given case, the agent resolved the case identifier and invoked the corresponding tools to retrieve their outputs. Every tool returned the pre-computed, validated readouts of its module, so that no quantitative value in the report was produced by the language model itself. The retrieved outputs were assembled into a report of predefined sections, each
corresponding to one module. Text in the report content was generated in two stages. First, for every
section, the model produced a concise clinical interpretation of each individual readout value
together with a short section-level summary, conditioned on the module values and on
established clinical decision thresholds. The model was constrained to restate only the
provided quantities and not to introduce values of its own. Second, a separate call
generated the overall assessment, synthesizing the assembled section content, the
readouts, their per-readout interpretations, and the section summaries, into a case-level
diagnosis and a concise account of the resulting treatment considerations. This synthesis step was explicitly restricted to information already present in the sections, so that the overall assessment condensed rather than extended the module-derived findings.

Quantitative readouts were contextualized according to their type. Diagnostic and
therapeutic markers were interpreted against their established clinical cut-offs \citep{reck_pembrolizumab_2016, da_silveira_correa_challenges_2024}. Readouts
that lack a single accepted threshold, such as the tissue-composition, tumor microenvironment, and immune-infiltration metrics, were instead contextualized against a
reference distribution derived from an independent cohort of lung cancer cases. For each such metric, a value was placed within the reference distribution and, where a direction
of clinical favorability could be defined, assigned a discrete favorability level from
the corresponding cohort quantiles (low, mid, high). This placement was computed deterministically from the reference cohort rather than by the language model, and was additionally supplied to the interpretation step so that the generated text remained consistent with it. We have included the prompt that the agents use to fill the report template in Appendix \ref{app:report_prompt}.

\subsection{Molecular Analysis}

Molecular analysis was performed as part of the routine diagnostic work-flow. For molecular profiling, tumor-rich regions were identified and annotated by pathologists using light microscopy (Olympus BX46), with tumor cellularity assessment prior to DNA extraction. Five to twenty serial Formalin-fixed, paraffin-embedded (FFPE) sections ($5~\mu\mathrm{m}$ thickness) were prepared, and DNA extraction was performed semi-automatically using the Maxwell RSC FFPE Plus DNA Purification Kit (Custom, Promega) according to the manufacturer's protocol. DNA concentration was quantified using the Qubit HS DNA assay (Thermo Fisher Scientific).

Sequencing libraries were generated using the AmpliSeq for Illumina Cancer Hotspot nNGM Panel v3 (Illumina) with 80 ng genomic DNA input per sample. The panel covers hotspot regions of 53 cancer-associated genes: \textit{AKT1, ALK, APC, ATM, BAP1, BRAF, BRCA1, BRCA2, CHEK2, CTNNB1, CUL3, DPYD, EGFR, ERBB2, ESR1, FGFR1, FGFR2, FGFR3, FGFR4, GNA11, GNAQ, GNAS, HRAS, IDH1, IDH2, JAK2, KEAP1, KIT, KRAS, MAP2K1, MEN1, MET, MLH1, MSH2, MSH6, NF1, NFE2L2, NRAS, NTRK1, NTRK2, NTRK3, PALB2, PDGFRA, PDGFRB, PIK3CA, PMS2, PTEN, RB1, RET, ROS1, SMARCA4, STK11, TERT}, and \textit{TP53}. Target regions were amplified by PCR using a Biometra TOne thermal cycler (Analytik Jena, Jena, Germany), followed by next-generation sequencing on an NextSeq Sequencing System (Illumina). Sequencing reads were aligned to the human reference genome (hg19), and variant calling was performed using SEQUENCE Pilot Software version 5.4.0 (JSI Medical Systems GmbH, Ettenheim, Germany). Variants were filtered using a minimum allele frequency threshold of 5\%, and pathogenic or likely pathogenic variants were retained for downstream analysis.

\subsection{Expert-Panel Adjudicated Reference Standard} \label{sec:consensus_ref}

To establish the reference standard for the prospective clinical
validation, all five participating thoracic pathologists re-evaluated all
70 cases after a washout period of at least three months following
completion of the initial assessment of the final case. Each case was
jointly reviewed by the expert panel, and a final task-specific score and
clinical classification were reached by discussion.

Expert-panel adjudication was chosen over simple majority voting because
joint expert review can reduce the influence of individual observer
variability and allows discrepant assessments to be resolved through
case-level re-evaluation \citep{krause_grader_2018, allison_understanding_2014}. Majority
voting, by contrast, may preserve systematic estimation biases when
several observers assess a feature in the same direction. This is particularly relevant for tumor cellularity,
for which visual assessment shows substantial interobserver variability
and has been reported to overestimate tumor content in a relevant
proportion of cases \citep{smits_estimation_2014, mikubo_calculating_2020, lhermitte_adequately_2017}.
This observation was also consistent with our data. Within the reference-standard cohort itself, the adjudicated tumor cellularity was systematically lower than the mean of
the individual panel members (median $-10$~ppt; lower in 86\% of cases;
$p<10^{-9}$, Wilcoxon signed-rank test), and four of the five raters
scored predominantly above the adjudicated value, indicating a shared,
same-direction tendency among the individual raters that joint review
resolved but a simple majority vote would have retained.

During adjudication, the histopathological slides were reviewed together
with LUCAID-derived heatmaps. These overlays are purely qualitative
spatial visualizations of tissue and cell distributions; the panel was
not shown the corresponding numeric module readouts (for example, the
model-derived tumor cellularity percentage, PD-L1 TPS, or MET and TROP-2
H-scores) or LUCAID's categorical classifications. The overlays therefore
facilitated visual inspection by displaying tissue and cell distributions
with greater contrast than the underlying histopathological image,
thereby making semiquantitative features easier to
assess visually, without disclosing the model's predicted values and thus
limiting the risk of anchoring the adjudicated scores to LUCAID's output.
Previous studies have likewise shown that AI-supported visualization or
quantitative assistance can improve the consistency of tumor cellularity
assessment by pathologists \citep{sakamoto_narrative_2020, kiyuna_evaluating_2024,
gertych_tumor_2025}. Review of the heatmaps also allowed the panel to assess
the plausibility of the model-derived spatial classifications in their
histopathological context, providing an additional quality-control layer
before their use as visual support during adjudication. The heatmaps
served as an adjunct to pathological review rather than as an independent
reference; final scores and clinical classifications were determined by
the expert panel after joint evaluation and discussion.

The resulting adjudicated assessments constituted the expert-panel
adjudicated reference standard used for all clinical-validation analyses.

\subsection{Clinical Action Concordance} \label{sec:cac}

To evaluate the agreement of LUCAID-derived continuous diagnostic scores with
clinically relevant decision categories, we defined a clinical action
concordance index (CAC). Continuous predictions from LUCAID and pathologist
assessments were converted into clinically relevant categorical classifications
using established thresholds and compared with the corresponding reference-standard
classifications (see Immunohistochemical Analysis section for further details).
The following clinical decision categories were applied: molecular testing
adequacy based on tumor cellularity ($<10\%$ versus $\geq 10\%$), PD-L1
expression based on tumor proportion score (TPS; $<1\%$, $1\text{--}49\%$ and
$\geq 50\%$), and MET as well as membranous and cytoplasmic TROP-2 expression
based on H-score categories (negative/weak $[0\text{--}100)$; moderate
$[100\text{--}200)$ and high $[200\text{--}300]$). CAC was calculated as the proportion of
cases in which the categorical assessment of each rater matched the
corresponding reference-standard classification. In this analysis the LUCAID rater denotes the readouts of the individual diagnostic modules rather than the agent's free-text output, which was evaluated separately for groundedness and interpretation (see the Report Generation results). Agreement was evaluated independently for each clinical action category. Interobserver disagreement was defined per assessment as a non-unanimous clinical-action category among the five pathologists, evaluated over the cases scored for each task.

\subsection{Pathologic Regression Grading After Neoadjuvant Therapy}

Pathologic response to neoadjuvant therapy was assessed in 140 resected non-small cell lung cancer cases treated with neoadjuvant chemotherapy or chemoimmunotherapy at the Evangelische Lungenklinik Berlin-Buch between 2020 and 2026 (Figure~\ref{fig6:clinicopathological_applications}d–f), following the IASLC multidisciplinary recommendations for pathologic assessment of lung cancer resection specimens following neoadjuvant therapy \citep{travis_iaslc_2020}. Two pathologists jointly reviewed all H\&E-stained slides of the tumor bed and estimated the percentages of (1) viable tumor, (2) necrosis, and (3) stroma, the latter comprising both fibrosis and inflammation, with the three components summing to 100\% of the tumor bed. In accordance with these recommendations, each component was recorded in 10\% increments, except for amounts of 5\% or less, which were recorded in single-percentage steps. For automated assessment, LUCAID quantified the same three components as area fractions of the segmented carcinoma, stromal, and necrotic compartments within the tumor region derived from the tissue segmentation module. Agreement between model-derived and pathologist estimates was assessed per component using Pearson and Spearman correlation coefficients (two-tailed, $\alpha$ = 0.05), and quantitative deviation was summarized as the mean absolute error of the model against the joint pathologist assessment.

\subsection{Spatial Feature and Survival Analysis}

Spatial tumor-microenvironment features were computed per tissue sample from AI-phenotyped cells, restricted to the seven analysis phenotypes (carcinoma, endothelial, fibroblast, granulocyte, lymphocyte, macrophage, and plasma cells) and averaged across tissue samples to the patient level. Each feature was related to overall survival in its own Cox proportional-hazards model together with UICC stage (stage-adjusted, not co-fitted), reporting the hazard ratio, 95\% confidence interval, the two-sided stage-adjusted Wald p and the unadjusted log-rank p of the resulting strata; cases with a missing value for a given feature were excluded from that feature's Cox model and log-rank test (feature-specific n in Supplementary Table~\ref{tab:S2_forest_results}). Each feature was dichotomized at the median  -- or, for bimodal or strongly skewed markers, at a Gaussian-mixture threshold placed at the trough between the two populations (rule and cut per feature in Supplementary Table~\ref{tab:S2_forest_results}). The novel spatial features comprised three families: niche fractions, obtained by building a Delaunay neighbor graph over the cells, partitioning it into spatial communities by compartment-aware Leiden community detection, and typing each community by its cell-type composition — a community being lymphoid (tertiary-lymphoid-structure–like) when lymphocytes plus plasma cells reached $\geq 40\%$ of its cells—with the feature value defined as the fraction of analyzed cells assigned to that niche; adjacency fractions, the proportion of neighbor-graph edges directly connecting a given cell-type pair (e.g., carcinoma–plasma cell); and cell–cell distances, written A–B, the median over the A cells of the distance to the nearest anchoring B cell. The novel spatial candidates were drawn from a screen of 586 spatial metrics with Benjamini–Hochberg false-discovery-rate correction across the whole screen (24 metrics at q < 0.05; the full screen and shortlist are provided as Supplementary Table 4 and 5). From this screen, the seven novel features shown in Figure~\ref{fig6:clinicopathological_applications}i were selected for biological interpretability, coverage of the spatial-feature families and both risk directions, and pathologist supervision, retaining four features that remained significant after false-discovery-rate correction and three significant at the nominal level only; the per-panel Benjamini–Hochberg q values reported in Supplementary Table~\ref{tab:S2_forest_results} are computed after this selection and describe the displayed panel.

\subsection{Statistical Analysis}

Module performance for quality control, tissue segmentation, and cell
phenotyping was evaluated with F1 scores, reported per class and as macro
averages, comparing predictions to pathologist annotations at the pixel level
(segmentation) or cell level (classification); IHC expression-scoring models
were evaluated analogously with per-category and macro-averaged F1 on the
annotated cells of each slide.

Continuous predictions were compared with their reference standard (expert-panel
consensus or, for tumor cellularity, \textit{KRAS} VAF) using Pearson and
Spearman correlation coefficients (two-tailed, $\alpha = 0.05$) and mean absolute
error (MAE) with 95\% confidence intervals from 1{,}000 bootstrap iterations.
Agreement with the pathologist consensus is reported as Spearman $\rho$ and
agreement with the molecular \textit{KRAS} reference as Pearson $r$; for both
\textit{KRAS} analyses, cases with an estimated tumor fraction (2 $\times$ VAF)
exceeding 100\% were excluded. All prospective analyses were restricted to the
70-case cohort scored for at least one IHC marker, so that every task was
evaluated on the same patients.
Slides with fewer than 100 tumor cells were excluded from percentage-based
analyses. All analyses used Python (scipy.stats, numpy).

\subsection{Ethical Approval and Consent}

Ethical approval for this study was granted by the Ethics Committee of Charité -- University Medical Center Berlin (EA4/082/22), and all procedures adhered to the ethical principles for medical research of the Declaration of Helsinki. All patients provided written informed consent for the scientific use of their archived tissue and associated clinical data. Information on race, ethnicity and socioeconomic status was not collected as part of routine diagnostic documentation and was therefore unavailable for analysis.

\section*{Acknowledgements}

Marie-Lisa Eich is a participant in the BIH Charité Digital Clinician Scientist Program funded by the Charit\'e -- University Medical Center Berlin and the Berlin Institute of Health at Charité (BIH).
Marie-Lisa Eich and Simon Schallenberg were co-funded by the Berlin’s development bank (PROFIT grant: 10191964).
Mihnea P. Dragomir is a participant in the BIH Charité Clinician Scientist Program funded by the Charit\'e -- University Medical Center Berlin, and the Berlin Institute of Health at Charité (BIH).
This work was partly funded by the German Ministry for Education and Research (under refs 01IS14013A-E, 01GQ1115, 01GQ0850, 01IS18056A, 01IS18025A, 13GW0744D, and BIFOLD25B) and by  DFG. Furthermore, Klaus-Robert M\"uller was partly supported by the Institute of Information \& Communications Technology Planning \& Evaluation (IITP) grant funded by the Korea government (MSIT) (No. RS-2019-II190079, Artificial Intelligence Graduate School Program, Korea University) and grant funded by the Korea government (MSIT) (No. RS-2024-00457882, AI Research Hub Project).

\section*{Competing interests}
F.K., M.A., and K.R.M. are co-founders of Aignostics. F.K. is lead medical advisor and board member of Aignostics. K.R.M. is lead technical advisor to Aignostics. M.A. is CTO of Aignostics. S.S. is a part-time employee at Aignostics. DH is a member of the scientific advisory board at Aignostics. The other authors declare no conflict of interest.

\bibliography{references}
\newpage

\appendix
\section{Supplementary Tables And Supplementary Figures}

\setcounter{figure}{0}
\setcounter{table}{0}
\captionsetup[figure]{name=Supplementary Figure}
\captionsetup[table]{name=Supplementary Table}

\begin{figure}[!ht]
    \centering
    \includegraphics[width=\linewidth,height=0.6\textheight,keepaspectratio,trim={0 4.0cm 0 0}]{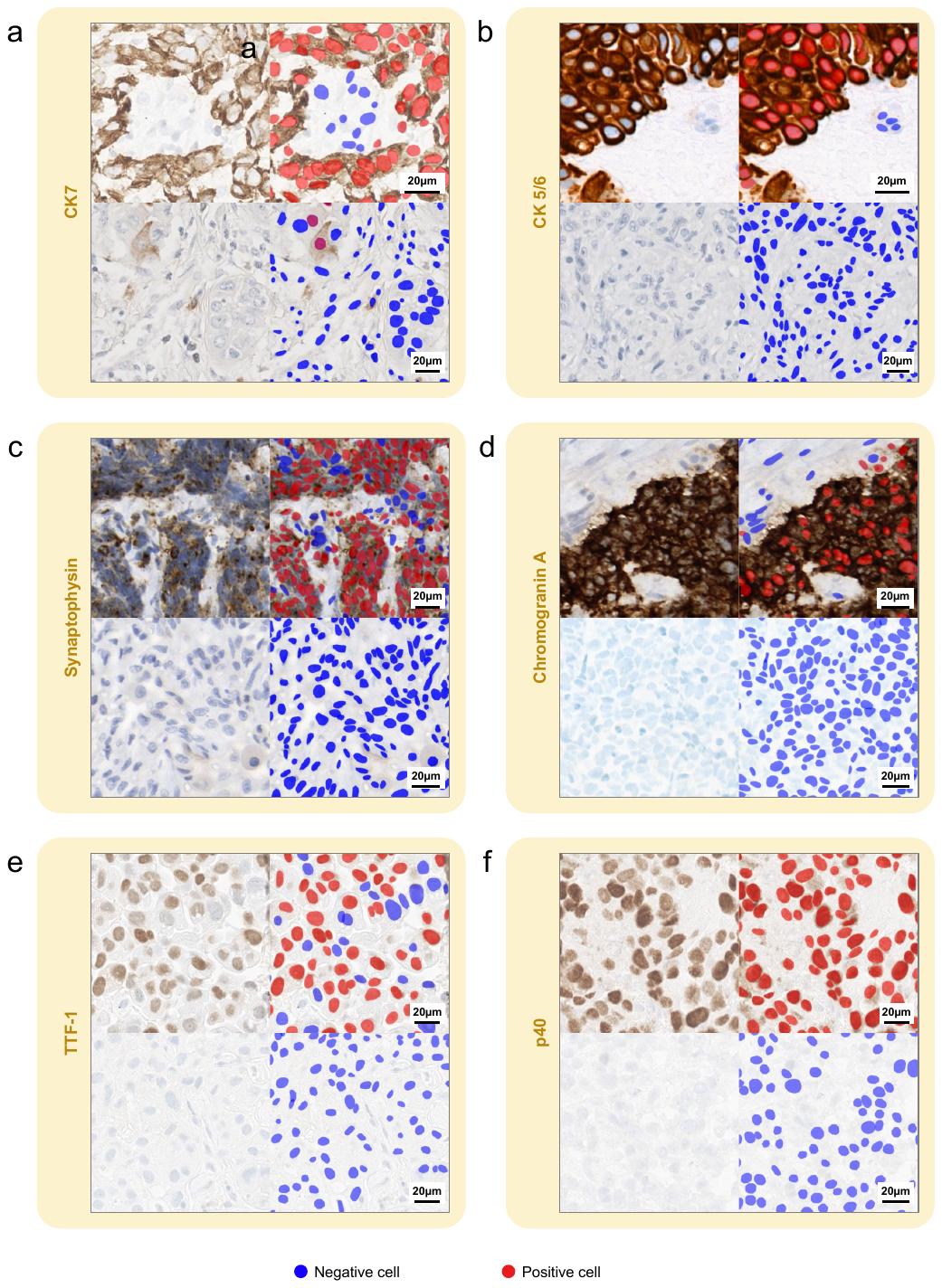}
    \small
    \caption[Immunohistochemical tumor
    typing.]{\textbf{Immunohistochemical Tumor Subtyping.}
    Representative examples of AI model expression scoring for the cytoplasmic markers
    \textbf{a}, CK7; \textbf{b}, CK5/6; \textbf{c}, synaptophysin; and \textbf{d}, chromogranin A; and for the nuclear markers \textbf{e}, TTF-1; and \textbf{f}, p40. For each marker, a positive case (upper two panels) and a negative case (lower two panels) are shown, each comprising the IHC image (left panels) and the corresponding AI model cell classification overlay (right panels). For cytoplasmic markers, positivity was determined based on LUCAID module 3 cytoplasmic biomarker expression scoring; for the nuclear markers TTF-1 and p40, positivity was determined by nuclear DAB intensity thresholding. Cases with more than 10\% positive tumor cells were classified as marker-positive.\\
    \textit{Abbreviations:} CK5/6: cytokeratin 5/6; CK7: cytokeratin 7;  DAS: 3,3'-Diaminobenzidin, IHC, immunohistochemical; TTF-1: thyroid transcription
    factor 1.}
    \label{sfig1:ihc_typing}
\end{figure}
\begin{figure}[p]
    \centering
    \includegraphics[width=\linewidth,height=0.7\textheight,keepaspectratio,trim={0 6.0cm 0 0}]{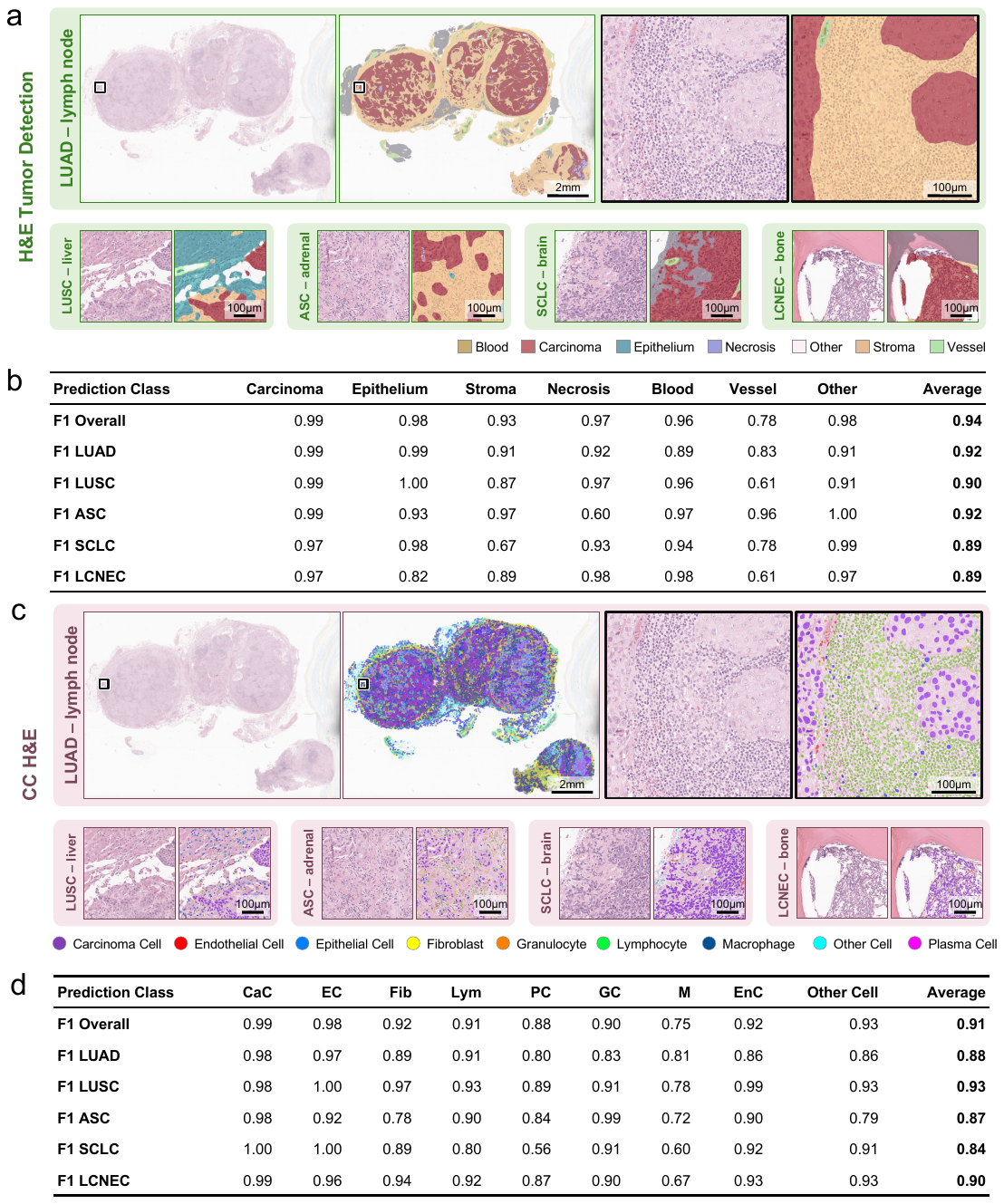}
    \small
    \caption[Correlation of AI Cell Classification with
    Molecular Analysis.]{\textbf{LUCAID H\&E Tumor Detection and Cell Phenotyping in Metastatic Lung Cancers}
\textbf{a}, H\&E tumor-detection examples in metastases from five major lung carcinoma subtypes, with
histopathological images shown on the left and corresponding model-derived tissue-segmentation overlays
on the right. For LUAD, a lymph node metastasis is shown as an overview image with a higher-magnification
view; the lower row shows metastases of LUSC (liver), ASC (adrenal gland), SCLC (brain) and LCNEC (bone)
from left to right. The segmentation model distinguishes seven tissue compartments: carcinoma, epithelium,
stroma, necrosis, blood, vessel and other. \textbf{b}, F1 scores for each tissue compartment overall and
stratified by histological subtype. \textbf{c}, H\&E cell-phenotyping examples in the same metastatic
specimens, with histopathological images shown on the left and corresponding model-derived
cell-classification overlays on the right. The model distinguishes nine cell classes: carcinoma cells,
endothelial cells, epithelial cells, fibroblasts, granulocytes, lymphocytes, macrophages, plasma cells and
other cells. \textbf{d}, F1 scores for each cell class overall and stratified by histological subtype.

Abbreviations: ASC, adenosquamous carcinoma; CaC, carcinoma cell; CC, cell classification; EC, endothelial
cell; EnC, epithelial cell; Fib, fibroblast; GC, granulocyte; H\&E, hematoxylin and eosin; LCNEC, large
cell neuroendocrine carcinoma; LUAD, lung adenocarcinoma; LUSC, lung squamous cell carcinoma; Lym,
lymphocyte; M, macrophage; PC, plasma cell; SCLC, small cell lung cancer.}
    \label{sfig2:metastasis}
\end{figure}

\begin{figure}[p]
    \centering
    \includegraphics[width=\linewidth,height=0.7\textheight,keepaspectratio,trim={0 0cm 0 0}]{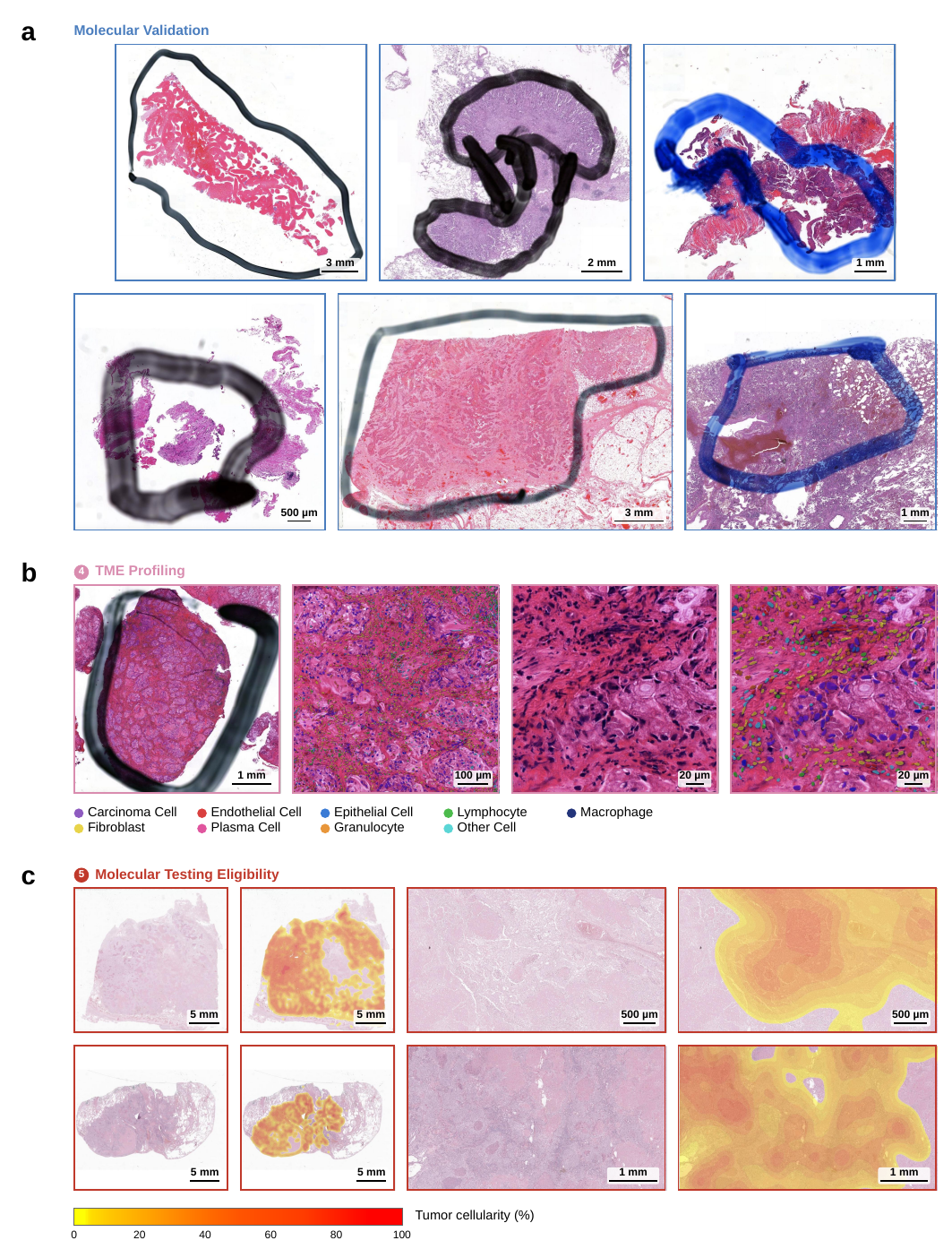}
    \small
    \caption[Correlation of AI Cell Classification with
    Molecular Analysis.]{\textbf{Molecular Validation of LUCAID Cell Phenotyping Module in Lung Cancer Patients.}
    \textbf{a}, Pathologist pen-marked H\&E whole slide images in six representative cases, each shown as a
whole-slide overview with the pen-marked region designating the area selected for molecular analysis.
\textbf{b}, TME profiling (module 4) in a representative lung adenocarcinoma. Whole-slide overview with the pen-marked tumor area (left), model overlay at intermediate magnification (second from left) and high magnification shown as original image and corresponding model overlay (two right panels). The model distinguishes nine cell classes: carcinoma cells, endothelial cells, epithelial cells, fibroblasts,
granulocytes, lymphocytes, macrophages, plasma cells and other cells. \textbf{c}, Molecular testing eligibility (module 5) in two representative cases (rows). Each case is shown as H\&E image and corresponding tumor cellularity heatmap at whole-slide overview (left pair) and high magnification (right pair). Color encodes predicted tumor cellularity (\%). 

\textit{Abbreviations:} H\&E, haematoxylin and eosin; TME, tumor microenvironment.}
    \label{sfig3:cellularity_molecular}
\end{figure}

\begin{figure}[p]
    \centering
    \includegraphics[width=\linewidth,height=0.7\textheight,keepaspectratio,trim={0 0cm 0 0}]{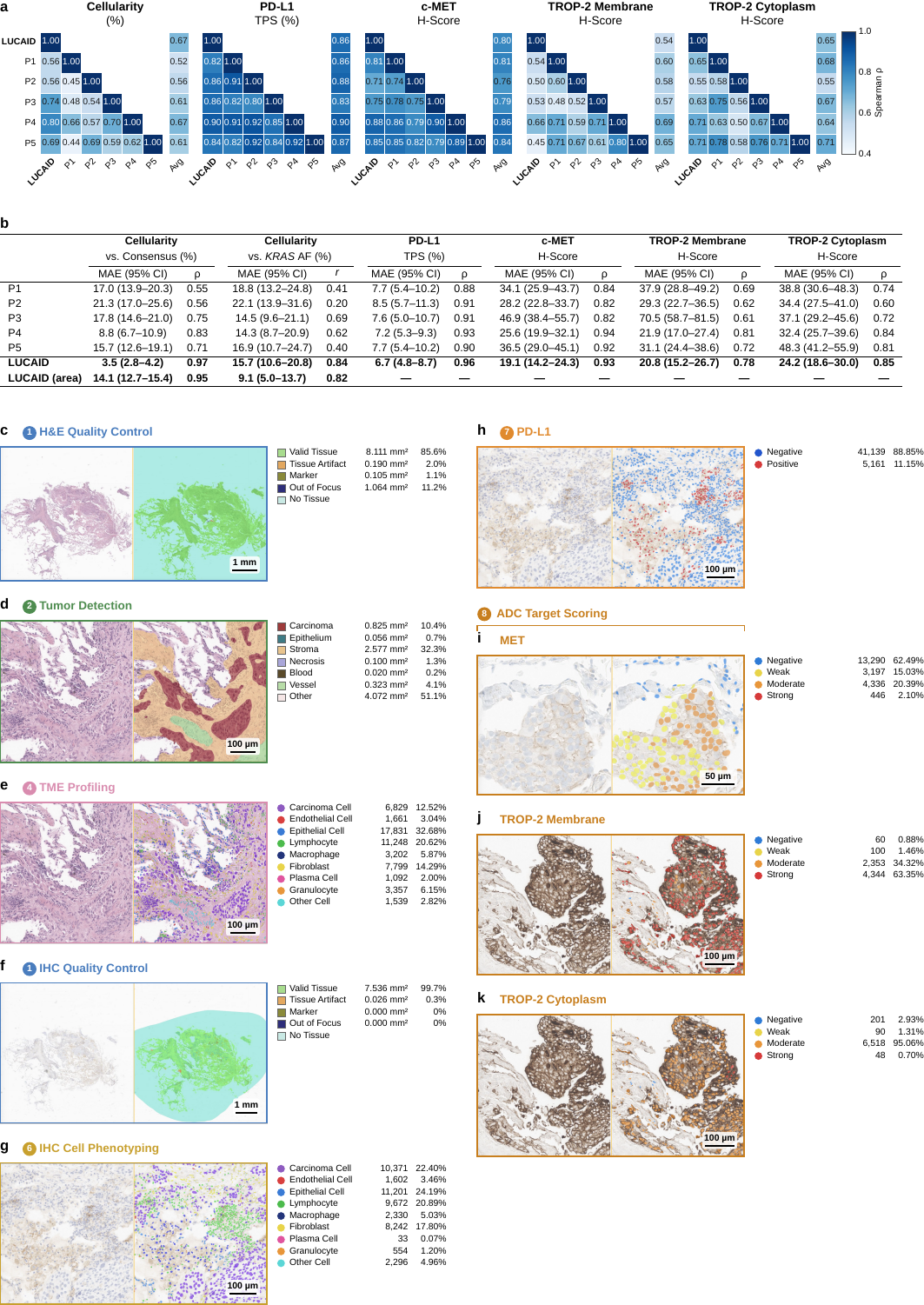}
    \small
    \caption[Clinical Validation of the AI Pipeline.]{%
    \textbf{Supplementary Figure 4: Extended prospective clinical validation across raters and LUCAID
modules.}
\textbf{a}, Pairwise Spearman correlation matrices for LUCAID and five pathologists (P1--P5) across tumor cellularity, PD-L1 tumor proportion score (TPS), MET H-score, TROP-2 membranous H-score and TROP-2 cytoplasmic H-score. Color encodes the Spearman correlation coefficient ($\rho$).
The rightmost column (Avg) shows each rater's mean pairwise correlation with all other raters. \textbf{b}, Mean absolute error (MAE) with 95\% confidence intervals and corresponding correlation coefficients (Spearman $\rho$, except Pearson $r$ for the \textit{KRAS} comparison) for LUCAID and each pathologist across tumor cellularity relative to the expert-panel adjudicated reference standard, tumor cellularity relative to \textit{KRAS} allele frequency (excluding cases with an estimated tumor fraction, 2 $\times$ VAF, $>$ 100\%), PD-L1 TPS,
MET H-score, TROP-2 membranous H-score and TROP-2 cytoplasmic H-score. LUCAID (area) denotes area-based rather than cell-based tumor cellularity estimation. \textbf{c--k}, Representative LUAD case showing sequential LUCAID module outputs, with histopathological images shown alongside the corresponding
model-derived overlays and the associated quantitative readouts. \textbf{c}, H\&E quality control.
\textbf{d}, Tumor detection. \textbf{e}, TME profiling. \textbf{f}, IHC quality control. \textbf{g}, IHC
cell phenotyping. \textbf{h}, PD-L1 scoring. \textbf{i--k}, ADC target scoring for MET (\textbf{i}),
TROP-2 membranous (\textbf{j}) and TROP-2 cytoplasmic (\textbf{k}) expression. 

\textit{Abbreviations:} ADC, antibody--drug conjugate; CI, confidence interval; H\&E, hematoxylin and
eosin; IHC, immunohistochemical; MAE, mean absolute error; MET, hepatocyte growth factor receptor; PD-L1,
programmed death-ligand 1; QC, quality control; TME, tumor microenvironment; TPS, tumor proportion score;
TROP-2, trophoblast cell-surface antigen 2.}
    \label{sfig4:corr}
\end{figure}

\begin{figure}[p]
    \centering
    \includegraphics[width=\linewidth,height=0.7\textheight,keepaspectratio,trim={0 0cm 0 0}]{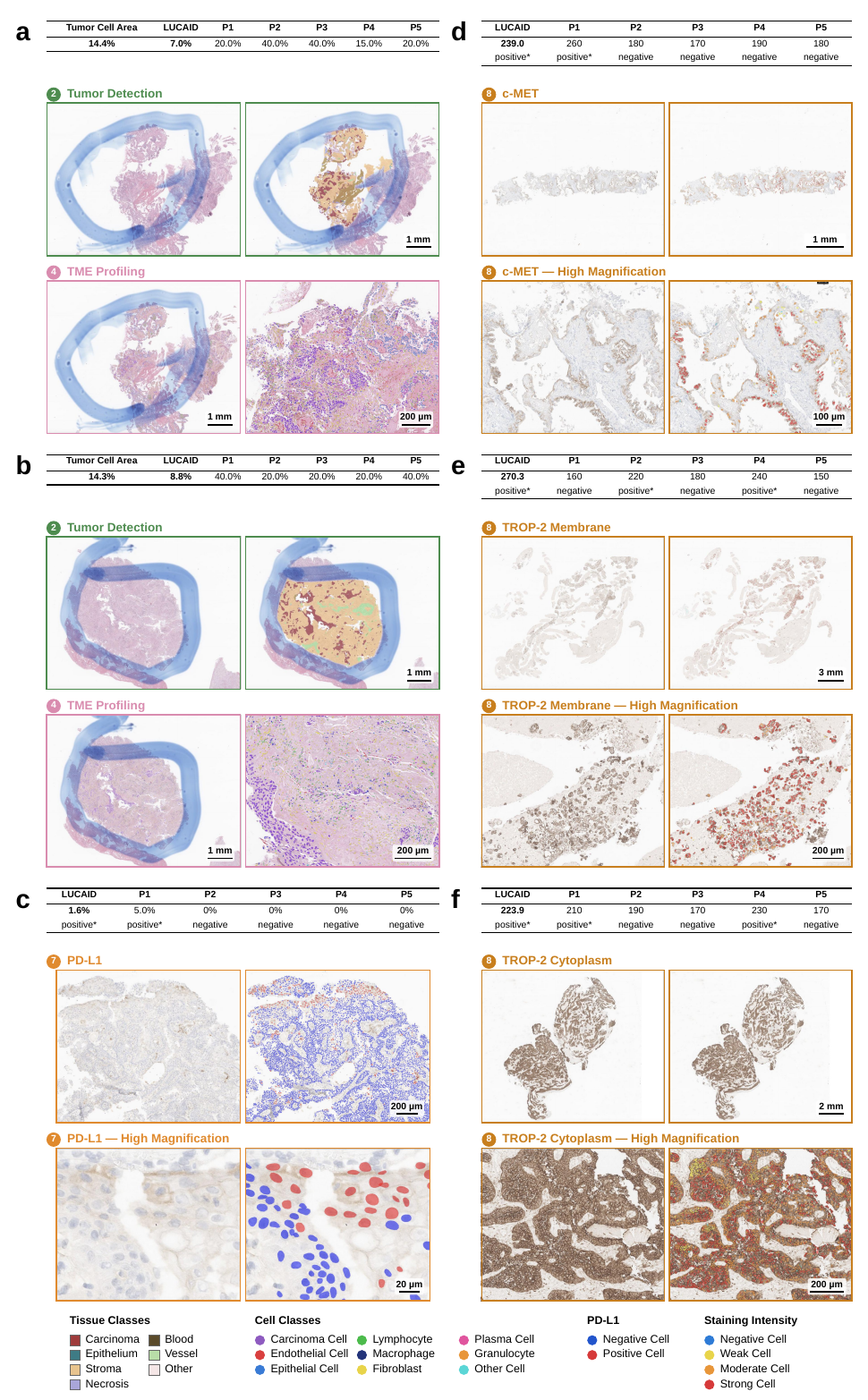}
    \small
    \caption[Diagnostic discordance between LUCAID and pathologists.]{\textbf{%
Representative discordant cases across tumor cell content estimation and IHC
biomarker scoring.} Tables report LUCAID and pathologist (P1--P5) scores;
asterisks denote scores above the clinical positivity threshold. Circled numbers
indicate the generating LUCAID module.
\textbf{a,b}, Tumor cell content. LUCAID estimated tumor cellularity at 7.0\%
(\textbf{a}) and 8.8\% (\textbf{b}), while all pathologists gave higher
estimates (15.0--40.0\%). Tumor Detection (module 2, upper) and TME Profiling
(module 4, lower) outputs are shown as histologic image (left) and model
overlay (right); high-magnification views show sparse carcinoma cells among
predominantly stromal and inflammatory components.
\textbf{c}, PD-L1 (module 7). LUCAID (TPS 1.6\%) and P1 (5.0\%) scored above the
1\% threshold; P2--P5 scored 0\%.
\textbf{d}, c-MET (module 8). LUCAID (H-score 239.0) and P1 (260) scored above
the positivity threshold (H-score $\geq$ 200); P2--P5 scored 170--190.
\textbf{e}, TROP-2 membranous (module 8). LUCAID (270.3), P2 (220) and P4 (240)
scored positive; P1 (160), P3 (180) and P5 (150) negative.
\textbf{f}, TROP-2 cytoplasmic (module 8). LUCAID (223.9), P1 (210) and P4 (230)
scored positive; P2 (190), P3 (170) and P5 (170) negative.
\\
\textit{Abbreviations:} MET, hepatocyte growth factor receptor; P1--P5,
pathologists 1--5; PD-L1, programmed death-ligand 1; TME, tumor
microenvironment; TPS, tumor proportion score; TROP-2, tumor-associated calcium
signal transducer 2.}
    \label{sfig5:discordance}
\end{figure}

\begin{table}[h]
\centering
\begin{threeparttable}
\caption{Clinicopathological Characteristics of the Prospective Clinical Validation Cohort}
\label{tab:S1_cohort_characteristics}
\begin{tabular}{llc}
\toprule
\textbf{Characteristics} & & \textbf{Total n (\%)} \\
\midrule
All Patients & & 70 \\
\addlinespace
Median Age (years) & & 63 (range 39--84) \\
\addlinespace
Gender & Female & 33 (47.1\%) \\
 & Male & 37 (52.9\%) \\
\addlinespace
Subtype & Lung adenocarcinoma & 47 (67.1\%) \\
 & Lung squamous cell carcinoma & 20 (28.6\%) \\
 & Large cell neuroendocrine carcinoma & 1 (1.4\%) \\
 & NSCLC NOS\tnote{a} & 2 (2.9\%) \\
\addlinespace
Localization & Primary tumor & 39 (55.7\%) \\
 & Lymph node metastasis & 14 (20.0\%) \\
 & Distant metastasis & 17 (24.3\%) \\
\addlinespace
Specimen type & Biopsy & 48 (68.6\%) \\
 & Fine needle aspiration & 16 (22.9\%) \\
 & Resection & 6 (8.6\%) \\
\bottomrule
\end{tabular}
\begin{tablenotes}
\small
\item[a] NSCLC NOS, non-small cell lung carcinoma, not otherwise specified.
\end{tablenotes}
\end{threeparttable}
\end{table}
\clearpage

\begin{table}[htbp]
\centering
\caption{\textbf{Supplementary Table 2: Clinicopathological Characteristics of the Discovery Cohort.}}
\label{tab:supp2_cohort}
\small
\begin{tabular}{@{}lrrrr@{}}
\toprule
 & Total & AC (LUAD) & SCC (LUSC) & ASC \\
Parameter & $n = 1{,}001$ & $n = 581$ & $n = 402$ & $n = 18$ \\
\midrule
\multicolumn{5}{@{}l}{\textit{Sex}} \\
\quad Male & 616 & 308 & 300 & 8 \\
\quad Female & 385 & 273 & 102 & 10 \\
\addlinespace
\multicolumn{5}{@{}l}{\textit{Age, years}} \\
\quad Median (IQR) & 66 (59--73) & 65 (58--72) & 68 (61--74) & 74 (68--76) \\
\addlinespace
\multicolumn{5}{@{}l}{\textit{Smoking status}} \\
\quad Never-smoker & 67 & 53 & 12 & 2 \\
\quad Non-smoker & 315 & 184 & 125 & 6 \\
\quad Smoker & 327 & 198 & 123 & 6 \\
\quad Not recorded & 292 & 146 & 142 & 4 \\
\addlinespace
\multicolumn{5}{@{}l}{\textit{pT stage}} \\
\quad T0/Tx & 3 & 1 & 2 & 0 \\
\quad T1 & 391 & 259 & 127 & 5 \\
\quad T2 & 317 & 188 & 121 & 8 \\
\quad T3 & 180 & 85 & 94 & 1 \\
\quad T4 & 109 & 48 & 57 & 4 \\
\quad Not recorded & 1 & 0 & 1 & 0 \\
\addlinespace
\multicolumn{5}{@{}l}{\textit{pN stage}} \\
\quad N0 & 633 & 376 & 245 & 12 \\
\quad N1 & 186 & 79 & 105 & 2 \\
\quad N2 & 150 & 105 & 41 & 4 \\
\quad N3 & 11 & 7 & 4 & 0 \\
\quad Not recorded & 21 & 14 & 7 & 0 \\
\addlinespace
\multicolumn{5}{@{}l}{\textit{pM stage}} \\
\quad M0 & 914 & 512 & 385 & 17 \\
\quad M1a & 14 & 7 & 7 & 0 \\
\quad M1b & 69 & 59 & 9 & 1 \\
\quad M1c & 4 & 3 & 1 & 0 \\
\addlinespace
\multicolumn{5}{@{}l}{\textit{UICC8 stage}} \\
\quad I & 446 & 274 & 163 & 9 \\
\quad II & 226 & 108 & 115 & 3 \\
\quad III & 242 & 130 & 107 & 5 \\
\quad IV & 87 & 69 & 17 & 1 \\
\addlinespace
\multicolumn{5}{@{}l}{\textit{Grade}} \\
\quad G1 & 51 & 50 & 1 & 0 \\
\quad G2 & 527 & 309 & 210 & 8 \\
\quad G3 & 303 & 156 & 140 & 7 \\
\quad Not recorded & 120 & 66 & 51 & 3 \\
\addlinespace
\multicolumn{5}{@{}l}{\textit{Resection status}} \\
\quad R0 & 919 & 539 & 363 & 17 \\
\quad R1 & 81 & 42 & 38 & 1 \\
\quad R2 & 1 & 0 & 1 & 0 \\
\addlinespace
\multicolumn{5}{@{}l}{\textit{Overall survival}} \\
\quad Deaths & 528 & 272 & 241 & 15 \\
\quad Follow-up, months, median (reverse KM) & 73 & 73 & 75 & 76 \\
\bottomrule
\end{tabular}

\vspace{0.5em}
\raggedright
\footnotesize
\textit{Abbreviations:} AC, adenocarcinoma; ASC, adenosquamous carcinoma; IQR, interquartile range; KM,
Kaplan--Meier; LUAD, lung adenocarcinoma; LUSC, lung squamous cell carcinoma; SCC, squamous cell carcinoma;
UICC8, Union for International Cancer Control staging system, 8th edition.
\end{table}

\begin{landscape}
\begingroup
\scriptsize
\setlength{\tabcolsep}{3pt}
\renewcommand{\arraystretch}{1.15}
\begin{longtable}{p{2.6cm} p{1.6cm} p{5.2cm} p{1.3cm} c c c c l c c c}
\caption{Forest Results of AI-quantified Spatial Features with Feature Definitions} \label{tab:S2_forest_results} \\
\toprule
\textbf{feature} & \textbf{category} & \textbf{formula} & \textbf{unit} & \textbf{HR} & \textbf{CI\_95} & \textbf{p\_Wald\_stage\_adjusted} & \textbf{q\_value\_BH\_panel} & \textbf{split\_rule} & \textbf{split\_cutoff} & \textbf{n} & \textbf{events} \\
\midrule
\endfirsthead

\multicolumn{12}{c}{\textit{Supplementary Table 2 (continued)}} \\
\toprule
\textbf{feature} & \textbf{category} & \textbf{formula} & \textbf{unit} & \textbf{HR} & \textbf{CI\_95} & \textbf{p\_Wald\_stage\_adjusted} & \textbf{q\_value\_BH\_panel} & \textbf{split\_rule} & \textbf{split\_cutoff} & \textbf{n} & \textbf{events} \\
\midrule
\endhead

\midrule
\multicolumn{12}{r}{\textit{Continued on next page}} \\
\endfoot

\bottomrule
\endlastfoot

Carcinoma--plasma cell adjacency & Novel spatial & fraction of neighbour-graph (delaunay) edges linking carcinoma to plasma cells & fraction & 0.75 & 0.63--0.88 & $<$0.001 & 0.009 & Median & 0.0024 & 1001 & 528 \\
Lymphoid niche (TLS-like) & Novel spatial & fraction of cells in lymphoid-typed leiden communities (lymphocyte + plasma $\geq$ 40\%; tertiary-lymphoid-structure--like aggregate) & fraction & 0.75 & 0.63--0.89 & $<$0.001 & 0.009 & Gaussian mixture & 0.0531 & 1001 & 528 \\
Stromal TIL \% & Established & cell\_\allowbreak percentage\_\allowbreak lymphocyte\_\allowbreak stroma & \% & 0.75 & 0.63--0.89 & 0.001 & 0.009 & Median & 18.3725 & 1001 & 528 \\
Lympho/granulocyte ratio & Established & dens(lymphocyte)/\allowbreak dens(granulocyte) & ratio & 0.78 & 0.65--0.92 & 0.004 & 0.023 & Median & 3.7804 & 1001 & 528 \\
Plasma cells \% & Composition & plasma /\allowbreak  total cells & \% & 0.79 & 0.67--0.94 & 0.007 & 0.034 & Median & 2.4752 & 1001 & 528 \\
Stromal cell density & Established & sum stromal-cell densities & cells/mm\textsuperscript{2} & 0.80 & 0.67--0.95 & 0.010 & 0.046 & Median & 4841.8387 & 996 & 524 \\
Macrophage--carcinoma cell distance & Novel spatial & median distance (\textmu m) from each macrophage within the carcinoma compartment to its nearest carcinoma cell (larger = macrophage exclusion) & \textmu m & 0.80 & 0.67--0.95 & 0.012 & 0.048 & Median & 13.8624 & 984 & 515 \\
Immune infiltration score & Established & immune cells /\allowbreak  total cells & fraction & 0.81 & 0.69--0.97 & 0.018 & 0.067 & Median & 0.2823 & 1000 & 527 \\
Carcinoma delineation & Established & largest\_\allowbreak perimeter\_\allowbreak carcinoma /\allowbreak  absolute\_\allowbreak area\_\allowbreak carcinoma & mm\textsuperscript{-1} & 0.82 & 0.69--0.98 & 0.026 & 0.081 & Median & 0.0125 & 993 & 522 \\
Lymphocytes \% & Composition & lymphocyte /\allowbreak  total cells & \% & 0.83 & 0.70--0.99 & 0.037 & 0.095 & Median & 9.5832 & 1001 & 528 \\
Plasma cell density & Composition & dens(plasma) & cells/mm\textsuperscript{2} & 0.83 & 0.70--0.98 & 0.032 & 0.089 & Median & 109.7638 & 1000 & 527 \\
Carcinoma--lymphocyte adjacency & Novel spatial & fraction of neighbour-graph (delaunay) edges linking carcinoma to lymphocytes & fraction & 0.83 & 0.70--0.98 & 0.031 & 0.089 & Median & 0.0197 & 1001 & 528 \\
Intratumoral TIL \% & Established & cell\_\allowbreak percentage\_\allowbreak lymphocyte\_\allowbreak carcinoma & \% & 0.84 & 0.71--1.00 & 0.052 & 0.127 & Median & 1.672 & 985 & 518 \\
Lympho/macrophage ratio & Established & dens(lymphocyte)/\allowbreak dens(macrophage) & ratio & 0.85 & 0.72--1.01 & 0.062 & 0.144 & Median & 1.2653 & 1001 & 528 \\
Macrophage density & Established & dens(macrophage) & cells/mm\textsuperscript{2} & 0.86 & 0.72--1.02 & 0.079 & 0.174 & Median & 381.45 & 1001 & 528 \\
Granulocyte density & Established & dens(granulocyte) & cells/mm\textsuperscript{2} & 0.87 & 0.73--1.03 & 0.116 & 0.243 & Median & 121.645 & 1001 & 528 \\
Lymphocytes/mm\textsuperscript{2} (whole tumor) & Established & dens(lymphocyte) & cells/mm\textsuperscript{2} & 0.88 & 0.74--1.05 & 0.152 & 0.288 & Median & 400.95 & 1001 & 528 \\
TIL/tumor ratio & Established & dens(lymphocyte)/\allowbreak dens(carcinoma) & ratio & 0.88 & 0.74--1.05 & 0.157 & 0.288 & Median & 0.3357 & 999 & 527 \\
Tumor area \% & Established & relative\_\allowbreak area\_\allowbreak carcinoma & \% & 0.89 & 0.75--1.06 & 0.190 & 0.334 & Median & 37.9113 & 1001 & 528 \\
CAF/tumor ratio & Established & dens(fibroblast in carcinoma)/\allowbreak dens(carcinoma) & ratio & 0.90 & 0.76--1.07 & 0.228 & 0.386 & Median & 0.0724 & 979 & 513 \\
Macrophages \% & Composition & macrophage /\allowbreak  total cells & \% & 0.91 & 0.77--1.08 & 0.294 & 0.431 & Median & 8.6485 & 1001 & 528 \\
Total immune density & Established & sum immune-cell densities & cells/mm\textsuperscript{2} & 0.91 & 0.77--1.08 & 0.283 & 0.430 & Median & 1191.8332 & 1000 & 527 \\
Epithelial \% & Composition & relative\_\allowbreak area\_\allowbreak epithelial\_\allowbreak tissue (non-tumour epithelium /\allowbreak  tissue area) & \% & 0.91 & 0.76--1.08 & 0.262 & 0.412 & Gaussian mixture & 0.0134 & 1001 & 528 \\
Tumor--stroma ratio & Established & carcinoma/\allowbreak (carcinoma+stroma) area & fraction & 0.92 & 0.78--1.09 & 0.349 & 0.463 & Median & 0.4818 & 1001 & 528 \\
Fibroblast density & Established & dens(fibroblast) & cells/mm\textsuperscript{2} & 0.92 & 0.78--1.09 & 0.358 & 0.463 & Median & 770.3 & 1001 & 528 \\
Macrophage/tumor & Established & dens(macrophage)/\allowbreak dens(carcinoma) & ratio & 0.93 & 0.78--1.10 & 0.403 & 0.501 & Median & 0.2987 & 998 & 526 \\
Immune exclusion index & Established & dens(lymph in stroma)/\allowbreak dens(lymph in carcinoma) & ratio & 0.93 & 0.78--1.11 & 0.410 & 0.501 & Median & 12.4736 & 983 & 516 \\
Normal tissue \% & Established & (epithelial+other+vessel)/\allowbreak tissue area & \% & 0.94 & 0.79--1.12 & 0.491 & 0.584 & Gaussian mixture & 4.2488 & 1001 & 528 \\
Granulocyte/tumor & Established & dens(granulocyte)/\allowbreak dens(carcinoma) & ratio & 0.95 & 0.80--1.13 & 0.551 & 0.638 & Median & 0.1149 & 998 & 526 \\
Stroma area \% & Established & relative\_\allowbreak area\_\allowbreak stroma & \% & 0.95 & 0.80--1.13 & 0.581 & 0.641 & Median & 39.2449 & 1001 & 528 \\
Granulocytes \% & Composition & granulocyte /\allowbreak  total cells & \% & 0.95 & 0.80--1.13 & 0.583 & 0.641 & Median & 2.8856 & 1001 & 528 \\
Vascularization index & Established & dens(endothelial) & cells/mm\textsuperscript{2} & 0.96 & 0.81--1.14 & 0.652 & 0.700 & Median & 68.8975 & 1001 & 528 \\
Vessel \% & Composition & relative\_\allowbreak area\_\allowbreak vessel (vessel /\allowbreak  tissue area) & \% & 0.98 & 0.82--1.16 & 0.804 & 0.804 & Median & 1.245 & 1001 & 528 \\
Fibroblasts \% & Composition & fibroblast /\allowbreak  total cells & \% & 1.02 & 0.86--1.21 & 0.794 & 0.804 & Median & 18.833 & 1001 & 528 \\
Blood \% & Composition & relative\_\allowbreak area\_\allowbreak blood (blood /\allowbreak  tissue area) & \% & 1.04 & 0.87--1.23 & 0.696 & 0.729 & Gaussian mixture & 0.6527 & 1001 & 528 \\
Endothelial \% & Composition & endothelial /\allowbreak  total cells & \% & 1.08 & 0.91--1.29 & 0.357 & 0.463 & Median & 1.5679 & 1001 & 528 \\
Tumor cell density & Established & dens(carcinoma) & cells/mm\textsuperscript{2} & 1.09 & 0.92--1.30 & 0.311 & 0.441 & Median & 1537.925 & 999 & 527 \\
Carcinoma \% & Composition & carcinoma /\allowbreak  total cells & \% & 1.11 & 0.93--1.31 & 0.240 & 0.392 & Median & 37.8114 & 1001 & 528 \\
Interface immunity (20\textmu m) & Established & interface lymphocyte density /\allowbreak  bulk lymphocyte density & ratio & 1.15 & 0.96--1.36 & 0.125 & 0.250 & Median & 1.0763 & 975 & 511 \\
Higher carcinoma cell--lymphocyte distance & Novel spatial & median distance (\textmu m) from each carcinoma cell within the carcinoma compartment to its nearest lymphocyte (larger = immune exclusion) & \textmu m & 1.22 & 1.03--1.45 & 0.024 & 0.081 & Median & 48.0712 & 993 & 522 \\
Necrosis \% & Composition & relative\_\allowbreak area\_\allowbreak necrosis (necrosis /\allowbreak  segmented tissue area) & \% & 1.27 & 1.07--1.52 & 0.006 & 0.034 & Gaussian mixture & 2.5339 & 1001 & 528 \\
Higher lymphocyte dispersion & Novel spatial & median nearest-neighbour distance (\textmu m) between lymphocytes (larger = scattered, not clustered) & \textmu m & 1.33 & 1.12--1.58 & 0.001 & 0.009 & Median & 13.1388 & 1001 & 528 \\
Higher endothelial cell--lymphocyte distance & Novel spatial & median distance (\textmu m) from each endothelial cell to its nearest lymphocyte (larger = immune excluded from vasculature) & \textmu m & 1.33 & 1.12--1.58 & 0.001 & 0.009 & Median & 23.9685 & 1001 & 528 \\
Tissue NLR & Established & dens(granulocyte)/\allowbreak dens(lymphocyte) & ratio & 1.37 & 1.16--1.63 & $<$0.001 & 0.009 & Median & 0.3378 & 1001 & 528 \\
UICC stage (per step) & Clinical anchor (UICC) & --- & --- & 1.42 & 1.31--1.54 & $<$0.001 & --- & --- (per stage step) & --- & 1001 & 528 \\

\end{longtable}
\endgroup
\end{landscape}

\section{Report Evaluation}
\label{app:rep_eval}

\framedpdfpage{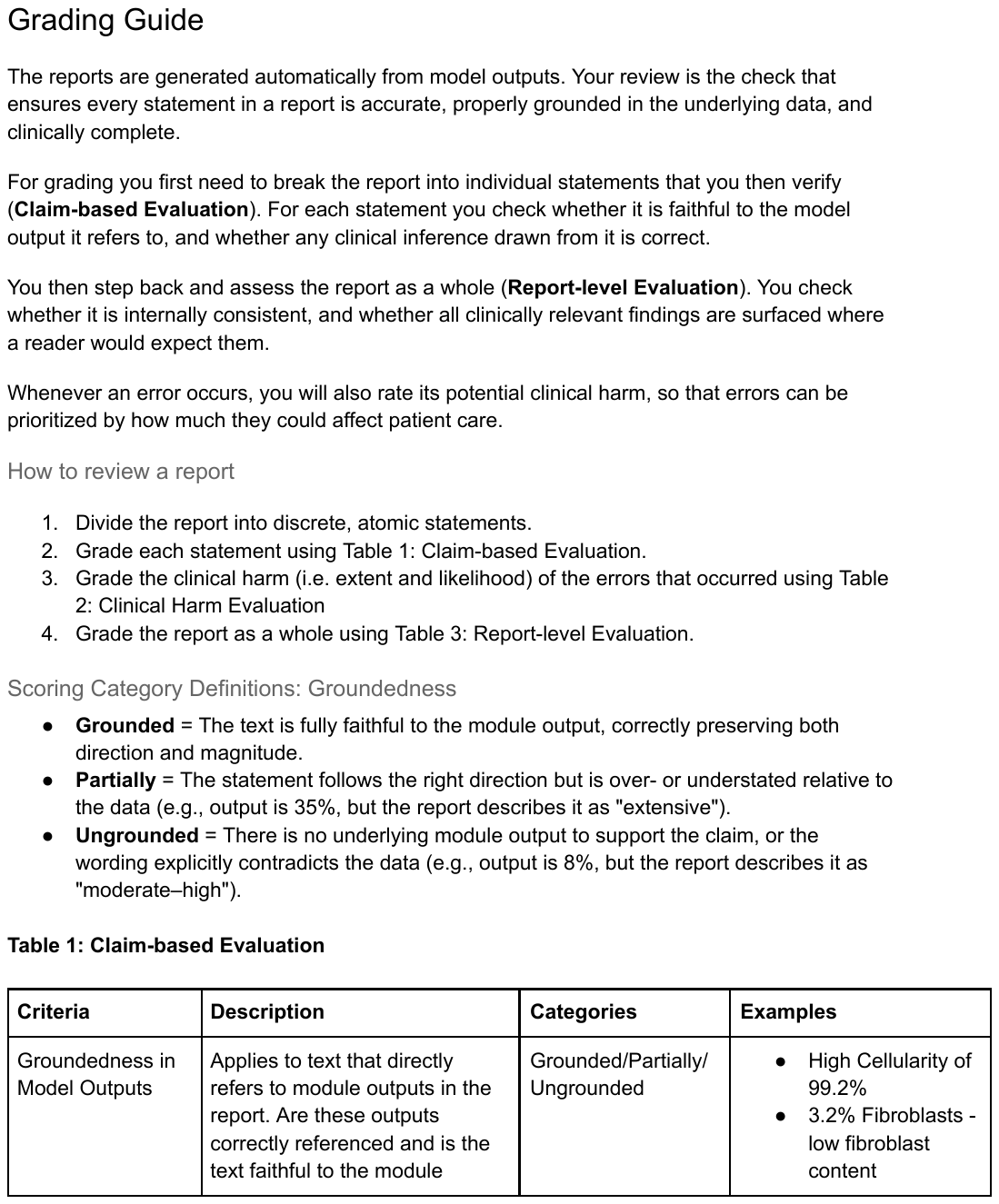}{1}
\foreach \p in {2,...,3}{%
  \framedpdfpage{Report_Evaluation_Framework_With_References.pdf}{\p}}

\section{Report Prompt} \label{app:report_prompt}
\begin{Verbatim}
You are a clinical pathology expert specializing in lung cancer diagnostics.
All cases are primary lung carcinomas.

Diagnostic Marker Section Content:
Interpret the provided markers and suggest a possible diagnosis based on the typical marker profiles of lung cancer subtypes. Do not provide any therapeutic interpretation or molecular testing recommendations in this section. Focus solely on subtype classification based on the marker data. For diagnostic markers, report each marker simply as positive or negative. Do NOT use the terms "TPS" or "H-score" and do not name the scoring method anywhere in this section's row interpretations or summary.

Diagnostic Marker Section context:
All marker values above 10% are positive. ADC: CK7 positive + p40 negative. When TTF1 negative then report TTF1 negative ADC, otherwise report TTF1 positive ADC. LCNEC: CK7 negative + synaptophysin/chromogranin A positive.
SqCC: CK5/6 positive + p40 positive + TTF1 negative

Prognostic Marker Section Content:
Interpret the provided prognostics markers in the context of therapeutic implications for lung cancer treatment. Do not make a subtype classification at this stage. Keep your assessment general about therapeutic implication based on the marker values and known thresholds.

Prognostic Marker Section context:
PD-L1 = [50-100%] -> recommend immune as first line therapy. (Monotherapy)
PD-L1 >0 until 50% -> Immune + chemotherapy. Suggest a specific immune checkpoint inhibitor if possible (e.g. pembrolizumab, atezolizumab). cMET: H-score: membranous H-score > 200 associated with improved response to cMET-directed ADCs (clinical trial context) Trop-2: membranous H-score >= 200 may be associated with improved response to TROP-2-directed ADCs (clinical trial context)

Molecular Panel Section Content:
For each row: one short sentence stating whether the alteration is actionable and any key threshold.

Molecular Panel Section context:
The molecular panel is the nNGM sequencing panel. Report only genes with detected mutations. Actionable alterations in lung cancer: KRAS G12C (sotorasib/adagrasib), EGFR exon 19 del / L858R (osimertinib),
BRAF V600E (dabrafenib+trametinib), ALK/ROS1/RET fusions (targeted inhibitors), MET exon 14 skip (capmatinib/tepotinib),
ERBB2 (trastuzumab deruxtecan), NTRK fusions (larotrectinib/entrectinib).
TP53 and KEAP1 are frequently mutated but not directly actionable; note them briefly. For the allele frequency: >20% is clonal (likely driver), <10% may be subclonal or artefact.

Section summary: 1-2 sentences on the most actionable finding(s) and molecular testing implications.

Guidelines:
- Use American English spelling throughout (tumor not tumour, vascularization, favor, characterize).
- For each row: write ONE short sentence (roughly 15 words or fewer) stating the key clinical fact.
  Do NOT use dashes (no em dash, no en dash, and no spaced hyphen used as a connector) and do NOT use
  semicolons. Commas are fine. Keep marker names intact (PD-L1, TTF-1, CK5/6, H-score).
  Reference thresholds where critical (PD-L1 >= 50%, MET H-score >= 200, TROP-2 eligibility).
  Example: "Expression is below the threshold for first-line pembrolizumab."
- For the section summary: 1-2 sentences maximum. Synthesize only the most actionable findings.
- Be factual. Do not fabricate data.
- For each row assign a traffic-light color (CSS hex string, or null) reflecting the PROGNOSTIC /
  PREDICTIVE value of the finding for the patient. The rule depends on the type of metric:

  Therapeutic / predictive markers (PD-L1, MET, TROP-2), GREEN or AMBER only, NEVER red:
    "#22a06b"  green: above threshold, a therapeutic option is indicated
                        (PD-L1 >= 50%, MET H-score >= 200, TROP-2 H-score >= 200). Opportunity wins:
                        color green even if high expression carries a worse baseline prognosis.
    "#e6a817"  amber: below threshold, no therapeutic option indicated (this is neutral, not red).

  Tissue Composition / TME Analysis / Immune Infiltration rows, do NOT choose a color:
    Their color is assigned deterministically in code from a 1001-case reference cohort and will
    OVERRIDE whatever you return, so return null for these rows. Each such row is annotated with its
    cohort position in brackets, e.g. "[cohort P72, high tertile, favorable end]" or
    "[cohort P50, middle tertile, descriptive]".
    Write in the register of a pathology report: ONE clinical sentence, roughly 20 to 30 words,
    covering: (1) briefly what the measure captures, (2) what it generally indicates prognostically in
    NSCLC, use the annotation's stated general direction ("higher values generally favorable" or
    "higher values generally unfavorable"), and state this for EVERY directional row INCLUDING ones
    where this case is intermediate, and (3) a qualitative read of THIS case's value (for example high,
    low, intermediate, sparse, abundant). Where it fits naturally and briefly, you MAY add a short
    clause on WHY the direction holds (a one-clause mechanism), but do not force it.
    Example: "Intratumoral TILs reflect cytotoxic lymphocytes engaging the tumor, and higher levels
    generally favor outcome because they signal active antitumor immunity, abundant in this case."
    Do NOT state the cohort tertile or percentile position in words (avoid "upper tertile", "low
    tertile", "mid cohort", "top of the cohort", "Nth percentile"): the bar already shows where the
    value sits. Do not use the word "descriptive" in the text.
    For metrics whose annotation says 'no established prognostic direction', give only what the measure
    captures and a qualitative read of the value, with NO favorable or unfavorable claim. Omit any prognostic comment.
    Vary the wording across rows so they do not read as a filled-in template. In particular, do NOT
    reuse one fixed phrase for the prognostic consequence: rotate how you express it rather than ending
    most rows with "a generally favorable/unfavorable feature in NSCLC". Draw on varied constructions
    such as "tends to portend better outcome", "usually carries a poorer prognosis", "a favorable
    prognostic sign", "often linked to reduced survival", "generally adverse in lung cancer" (do not
    reuse a single one). Stay consistent with the annotation (a value flagged low must read as low, not
    high). Do not copy the bracket punctuation, and follow the no-dashes and no-semicolons rule above.
    Rows with no annotation carry no cohort reference, so state the finding plainly.

  Cellularity (sample-adequacy metric):
    green >= 20% (adequate for molecular testing), amber 10-19% (borderline), red < 10% (insufficient).

  null (grey), NOT APPLICABLE: the metric carries no prognostic/predictive meaning. Use null for
    diagnostic lineage markers (TTF-1, p40, CK7, CK5/6, Chromogranin A, Synaptophysin) and for purely
    descriptive counts. These render grey.

  Red must only appear for a genuinely poor prognostic outcome or insufficient cellularity, never
  merely because a targeted therapy is unavailable.
\end{Verbatim}

\end{document}